\documentclass[lettersize,journal]{IEEEtran}
\usepackage{amsmath,amsfonts}
\usepackage{algorithmic}
\usepackage{algorithm}
\usepackage{array}
\usepackage[caption=false,font=normalsize,labelfont=sf,textfont=sf]{subfig}
\usepackage{textcomp}
\usepackage{stfloats}
\usepackage{url}
\usepackage{verbatim}
\usepackage{graphicx}
\usepackage{multirow}
\usepackage{textcomp}
\usepackage{pifont}
\usepackage{cite}
\usepackage{booktabs}
\usepackage{acronym}
\acrodef{HTR}{Handwritten Text Recognition}
\acrodef{MDLSTM}{Multi Dimensional LSTM}
\acrodef{SAAR}{\textit{Scan, Attend and Read}}
\acrodef{JLSAT}{\textit{Joint Line Segmentation and Transcription}}
\acrodef{VAN}{\textit{Vertical Attention Network}}
\acrodef{SPAN}{\textit{Simple Predict and Align Network}}
\acrodef{FPHR}{\textit{Full Page Handwritten Recognition model}}
\acrodef{DAN}{\textit{Document Attention Network}}
\acrodef{CER}{Character Error Rate}
\acrodef{WER}{Word Error Rate}
\acrodef{RO}{Reading Order}

\newtheorem{definition}{Definition}

\begin{document}

\title{Handwritten Text Recognition: A Survey}

\author{Carlos Garrido-Munoz, Antonio Rios-Vila, and Jorge Calvo-Zaragoza
% <-this % stops a space
\thanks{Authors are with the Pattern Recognition and Artificial Intelligence Group, University of Alicante, Spain (e-mails: carlos.garrido@ua.es, antonio.rios@ua.es, jorge.calvo@ua.es}% <-this % stops a space
% \thanks{Manuscript received XXXX; revised XXXX.}
}


% \IEEEpubid{0000--0000/00\$00.00~\copyright~2021 
% IEEE}

\maketitle

\begin{abstract} 
Handwritten Text Recognition (HTR) has become an essential field within pattern recognition and machine learning, with applications spanning historical document preservation to modern data entry and accessibility solutions. The complexity of HTR lies in the high variability of handwriting, which makes it challenging to develop robust recognition systems. This survey examines the evolution of HTR models, tracing their progression from early heuristic-based approaches to contemporary state-of-the-art neural models, which leverage deep learning techniques. The scope of the field has also expanded, with models initially capable of recognizing only word-level content progressing to recent end-to-end document-level approaches. Our paper categorizes existing work into two primary levels of recognition: (1) \emph{up to line-level}, encompassing word and line recognition, and (2) \emph{beyond line-level}, addressing paragraph- and document-level challenges. We provide a unified framework that examines research methodologies, recent advances in benchmarking, key datasets in the field, and a discussion of the results reported in the literature. Finally, we identify pressing research challenges and outline promising future directions, aiming to equip researchers and practitioners with a roadmap for advancing the field.
\end{abstract}


% === INTRODUCTION === 
\section{Introduction}
\label{sec:introduction}

\IEEEPARstart{H}{andwritten} Text Recognition (HTR) represents a cornerstone challenge within the fields of pattern recognition and machine learning. The task of converting handwritten text into a machine-encoded format has profound implications across several domains, including historical document preservation, automated data entry, digital note-taking, and accessibility for the visually impaired \cite{muehlberger2019transforming}. As we advance further into the digital age, the necessity for accurate and efficient HTR systems continues to grow, driven by the ever-increasing amount of handwritten data generated daily.

The complexity of HTR arises from the inherent variability in human handwriting. Unlike printed text, which adheres to standardized fonts and spacing, handwritten text exhibits a broad variety of styles, slants, sizes, and embellishments that are influenced by individual writing habits, cultural contexts, and even emotional states \cite{zakraoui2023study}. These variations present significant challenges for recognition systems, requiring algorithms capable of generalizing across diverse handwriting samples.

Historically, HTR systems have evolved through several technological waves. Early approaches relied on heuristic and rule-based methods, leveraging handcrafted features to recognize characters and words \cite{lexicon_kim_1997, offline_he_1994, hmm_guillevic_1997, handwritten_caesar_1994, offline_bunke_1994}. These systems were limited in their capacity to handle the high variability of handwritten text and often required significant preprocessing and segmentation of individual characters \cite{lu_segmentation_1996, casey_segmentation_1996}. The advent of machine learning introduced a paradigm shift, enabling the development of more adaptable and robust recognition systems. Statistical methods, such as Hidden Markov Models (HMM) \cite{hmm_guillevic_1997, hmmbased_elyacoubi_1999}, offered improved performance by learning from labeled data to recognize complete words or lines \cite{introduction_rabiner_1986, tutorial_rabiner_1989, offline_chen_1994, recognition_kundu_1989}.

The emergence of deep learning \cite{lecun2015deep} has revolutionized the field of HTR, bringing forth unprecedented advancements in recognition performance. Convolutional Neural Networks (CNN) \cite{Lecun98} and Recurrent Neural Networks (RNN) \cite{Hochreiter-LSTM} have demonstrated remarkable success in capturing spatial and sequential dependencies in handwriting. More recently, Transformer models have further pushed the boundaries \cite{TransformerVaswani}. These architectures have shown immense potential in HTR due to their ability to model long-range dependencies and handle variable-length sequences effectively.

\begin{figure}
    \centering
    \includegraphics[width=1\columnwidth]{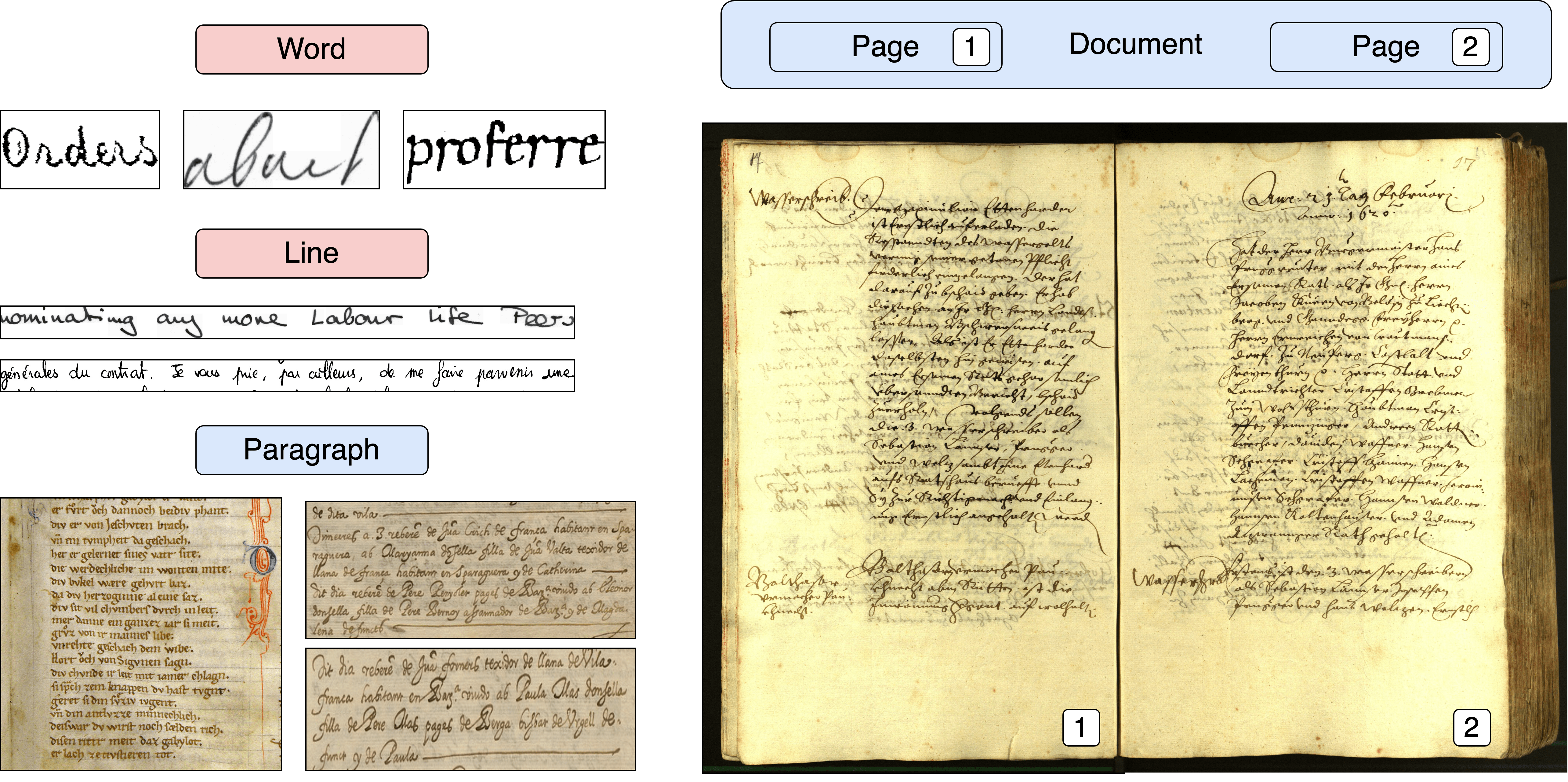}
    \caption{Overview of the different levels of granularity in Handwritten Text Recognition (HTR) systems, ranging from word-level to document-level transcription. The hierarchical structure illustrates the increasing complexity of HTR approaches as they seek to handle more extensive handwritten content, reflecting the advancements in this field.}
    \label{fig:htr-levels-overview}
\end{figure}

In addition to advances in performance, these technological breakthroughs have also diversified the approaches to HTR. The scope of recognition has expanded in complexity over time, from transcribing words and lines \cite{Graves2009_Line, multidimensional_recurrent_puig_2017, pay_attention_kang_2022, bluche_deep_2015} to approaches that directly output the content of entire paragraphs or documents \cite{Rouhou:PRL:2022, Coquenet:TPAMI:2023b, Bluche:NIPS:2016}. Within this context, this survey aims to provide a comprehensive examination of the current state of HTR focusing on its different levels of granularity. We broadly categorize the possibilities into two main groups (see Fig. \ref{fig:htr-levels-overview}): {\it up to line-level}, comprising word-level and line-level; and {\it beyond line-level}, involving paragraph-level and document-level recognition. The need for this distinction is evident in the different ways of approaching the associated challenges, which we will discuss later.

Our survey focuses on the transcription process itself. We will not delve into auxiliary processes such as character segmentation or layout analysis, which are often necessary to prepare the input for HTR systems. This distinction is crucial as it emphasizes methods that perform the handwriting recognition process. Therefore, when we discuss about \emph{up to line-level HTR}, we refer to approaches that directly process an image containing, at most, one line of text. Similarly, when we review paragraph- or document-level literature, we refer to approaches that directly transcribe the handwritten content at those levels, without relying on prior segmentation. Following this perspective, we purposely exclude character-level recognition from this survey, as it essentially reduces to a classification task, an area already well-covered in the vast existing literature. 

\begin{figure*}
    \centering
    \includegraphics[width=\linewidth]{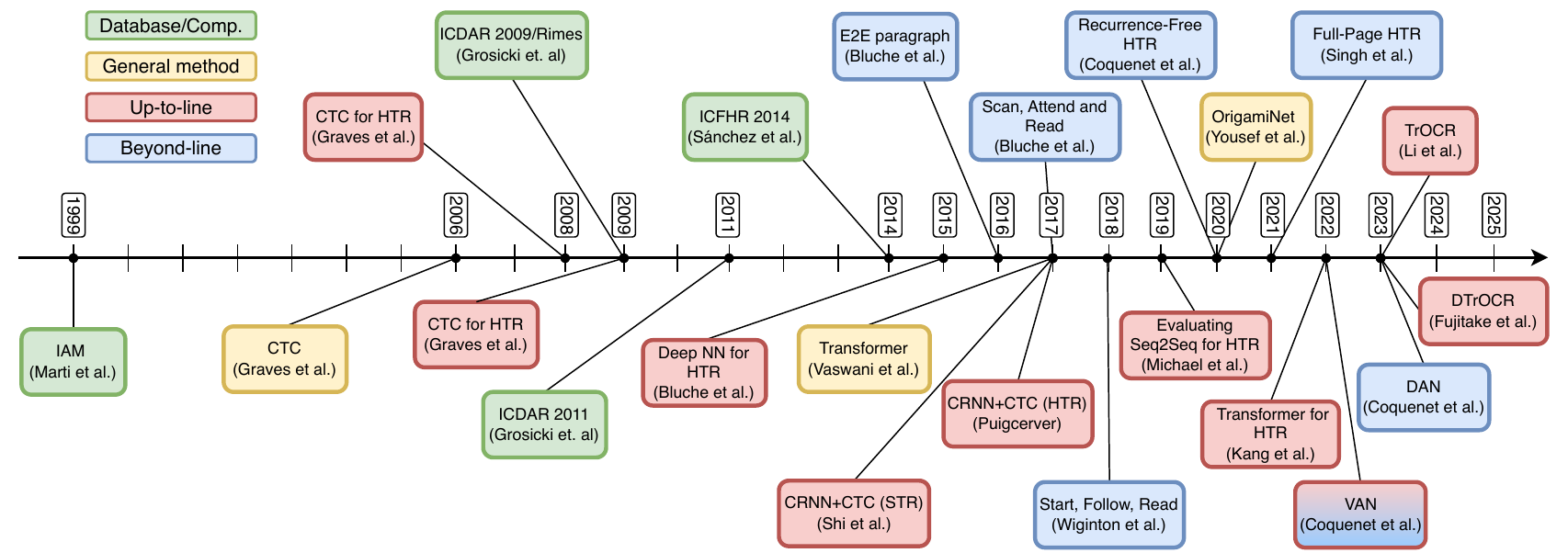}
    \caption{Timeline of milestones in Handwritten Text Recognition (HTR). We categorize them into four levels: datasets and competitions (green), general methods/architectures (yellow), up-to-line models (red), and beyond-line models (blue).}
    \label{fig:timeline-HTR}
\end{figure*}

\subsection{Purpose and contribution}
HTR has seen significant progress in recent years, driven by the development of sophisticated deep learning techniques and the growing availability of diverse datasets (see Fig. \ref{fig:timeline-HTR}). These advancements have transformed the field.

Existing surveys, such as those by Wang et al. \cite{wang2023survey} and AlKendi et al. \cite{alkendi2024advancements} provide comprehensive overviews of text detection and recognition across various contexts. Wang et al. cover a wide range of text recognition tasks, including printed and scene text, addressing both detection and recognition phases. AlKendi et al. \cite{alkendi2024advancements} offer a detailed examination of HTR systems with a specific focus on historical French documents and a broad range of languages. Additionally, Chen et al. \cite{chen2021text} focus on text recognition in the wild, dealing with text in natural scenes, which presents different challenges compared to HTR (as discussed later). Therefore, it is both timely and necessary to conduct a comprehensive survey of the current state of HTR that reflects the new landscape of the field. 

Our survey provides a specialized and focused examination of HTR, delving into the end-to-end transcription processes and categorizing methodologies based on the granularity of recognition. We provide a comprehensive review with a general and unifying view of the formulation. Our survey also explores the implications of these advancements on benchmarking practices, offering insights into the datasets and metrics most relevant for evaluating HTR systems. In addition, this work unifies and discusses the results reported by the most notable advances in the field through a common benchmark, providing perspective on the evolution of methodological impact in the field.

It is important to emphasize that this evolution analysis is framed within the transcription of Latin scripts, as most technological advancements in the field have been reported on these sources.

\subsection{Structure of the survey}
The structure of this survey is organized as follows: Section \ref{sec:preliminaries} covers the foundational preliminaries, including the historical background of HTR, the formal problem formulation, and its connections to related fields. Section \ref{sec:methods} explores the different methodologies employed in HTR, categorized by its inherent complexity levels, and examines the architectural designs and strategies used in each category. Section \ref{sec:benchmark} focuses on the benchmarking of HTR systems, providing an in-depth review of the datasets, evaluation metrics, and the performance reported in the literature. Finally, Section \ref{sec:conclusions} concludes the survey by summarizing the key insights and identifying open research challenges, outlining potential directions for future work in HTR technologies.

% ============================
% === BACKGROUND === 
% ============================
\section{Preliminaries}
\label{sec:preliminaries}

\subsection{Background of HTR}
\label{subsec:background}
In this section, we provide an overview of the foundational concepts and historical development of HTR, covering the shift from early heuristic-based approaches to modern deep learning techniques. This background sets the stage for a detailed exploration of the methodologies and advancements that have emerged in recent years. Initial attempts in HTR focused on classifying individual characters \cite{Lecun98}. However, within the scope of this survey, this task is considered a general classification challenge, that is extensively covered in the broader literature. Therefore, we will directly move on to the background of the sequential challenges of HTR.

Historically, HTR has adapted advancements from the field of Automatic Speech Recognition (ASR) \cite{tutorial_rabiner_1989, automatic_benzeghiba_2007, convolutional_abdelhamid_2014, statistical_jelinek_1997} because of the similarities in their problem formulations. Both fields deal with sequential data and require models that can handle variability and context dependencies. For years, the field of HTR relied on HMM \cite{neural_knerr_1998, hierarchical_dreuw_2011, offline_seiler_1996, offline_tay_2001}. Further advances favored the use of RNNs due to their capacity to model sequential data \cite{connectionist_graves_2006, empirical_chung_2014, recurrent_lai_2015, recurrent_hidasi_2020}.  In particular, the use of bidirectional Long Short-Term Memory (LSTM) networks with Connectionist Temporal Classification (CTC)~\cite{connectionist_graves_2006} dominated the state of the art in HTR for decades \cite{boosting_aradillas_2021, icdar2017_snchez_2017, icfhr2014_snchez_2014, international_abed_2010}. Despite this prevalence of CTC, attention-based encoder-decoder approaches \cite{neural_bahdanau_2015} gained popularity for their competitive results \cite{attentionhtr_kass_2022, lexicon_kumari_2022, attentionbased_abdallah_2020, endtoend_coquenet_2022}. The work of Michael et al. \cite{evaluating_michael_2019}, provides a comprehensive comparison of the different sequence-to-sequence approaches for line-level HTR.

As occurred in many other areas, there has been a growing interest in scalable and parallelizable architectures such as the Transformer \cite{TransformerVaswani} by adapting the Vision Transformer \cite{Dosovitskiy2020AnII} to the HTR field \cite{training_barrere_2024, weakly_paul_2023, improving_sang_2019, dtrocr_fujitake_2023, rethinking_diaz_2021, characterbased_poulos_2021}. HTR has benefited from this adaptation, either in isolation with an encoder-decoder \cite{trocr_li_2023, transformerbased_momeni_2023, ocformer_mostafa_2021, transformer_wick_2021} or in combination with the CTC objective function \cite{rescoring_wick_2021, dan_coquenet_2023, light_barrere_2022}. Diaz et al. \cite{rethinking_diaz_2021} explore universal architectures for text recognition, concluding that a CNN backbone with a Transformer encoder, a CTC-based decoder, plus an explicit language model, is the most effective strategy to date for line-level transcriptions. Regardless of this progress, however, the need for large labeled corpora as a pre-training strategy in Transformer-based models has become increasingly evident \cite{trocr_li_2023, rethinking_diaz_2021, vilbert_lu_2019, xlnet_yang_2019, simpler_lu_2021}. 

Given the increase in computational capacity, the paradigm has shifted towards end-to-end methods capable of reading beyond the line level. This current trend in the field began to develop with attention masking mechanisms that replace the explicit Layout Analysis (LA)~\cite{Bluche:NIPS:2016,Bluche:ICDAR:2017,Coquenet:TPAMI:2023}. Later advances also extended the single-line transcription state of the art to work with multiple lines. The CTC-based methodologies were extended to work in a multi-line setting, with reshaping mechanisms that alleviate its constraints~\cite{Yousef:CVPR:2020, Coquenet:ICDAR:2021}. The Transformer architecture also emerged in this paradigm, enabling the adaptation of attention-based encoder-decoder architectures to transcribe documents of any structure, thanks to its unconstrained nature and the ability to produce large synthetic datasets~\cite{Singh:ICDAR:2021,Coquenet:TPAMI:2023b,Dhiaf:PRL:2023}. This approach currently represents the state of the art of the field and lays the foundation for future development (see Fig. \ref{fig:timeline-HTR}).

\subsection{Levels of complexity}
The field of HTR encompasses a wide range of challenges that vary significantly in their complexity. While existing literature provides various perspectives on these challenges, it often lacks clear differentiation between input types and their inherent difficulties. To advance our understanding and properly evaluate different methodologies, establishing a precise framework that defines and categorizes the different levels of complexity is essential.

The task of HTR involves converting handwritten text into an encoded format that can be interpreted by a machine. However, the transcription challenge can be narrowed down in different ways. At the most fundamental level, we find individual \textit{characters},\footnote{We emphasize that ``character'' here refers to the basic graphical unit that an {HTR} model learns to recognize, as the concept may vary across different writing systems.} which represent the simplest unit of recognition and complexity. As stated above, character recognition is outside the scope of this paper, as it can be reduced to a classification problem.

%Therefore, 
Beyond character recognition, the concept of \ac{RO} becomes an important aspect to consider. The \ac{RO} is the sequence in which text elements are read to preserve the logical flow and conceptual meaning, following linguistic and cultural conventions. At the first level of complexity concerning \ac{RO}, we focus on \textit{words}, which are formed by the sequential arrangement of characters. For word-level recognition, the \ac{RO} follows a single and consistent direction, which is determined by the reading rules of the writing system. For example, in Latin sources it is left-to-right, while it is top-to-bottom in traditional Japanese and Chinese books.

\begin{figure}[h]
    \centering
    \includegraphics[scale=0.45]{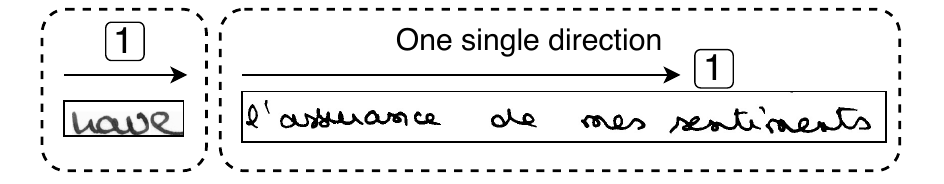}
    \caption{Reading Order (RO) with a single direction at the line level.}
    \label{fig:ro_one_direction}
\end{figure}

At a similar level of complexity, we also encounter lines, which are understood as a sequence of words separated by a delimiter character (commonly a space: ` '). This increase in complexity does not alter the \ac{RO}; instead, it lies in the need to transcribe a greater number of characters while incorporating this additional delimiter. In this context, HTR involves processing images that might span the full width of the page. These lines are read following the same \ac{RO} type as words, remaining consistent and unidirectional, as illustrated in Fig. \ref{fig:ro_one_direction}. Therefore, we establish the first level of complexity in the field.

\begin{definition}
Line-level HTR is the process of transcribing handwritten text from an input image containing a single line of text, which must be interpreted following a single \ac{RO}, as determined by the rules of the language in which it is encoded. This level encompasses both \textit{words} and \textit{lines}.
\end{definition}

\begin{figure}[h]
    \centering
    \includegraphics[scale=0.45]{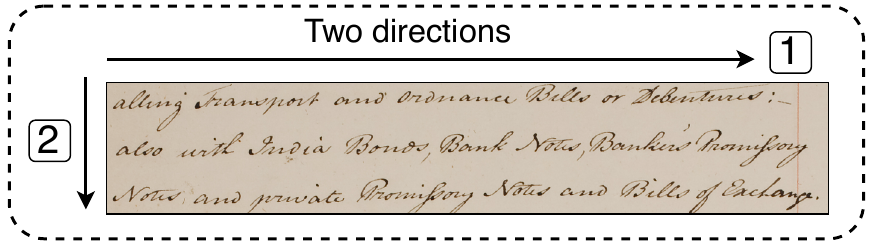}
    \caption{Reading Order (RO) with two directions at the paragraph level.}
    \label{fig:ro_two_directions}
\end{figure}

Given the spatial limitations of paper, it is unfeasible to write complete texts in a single line. For this reason, the stacking of lines in an additional dimension is traditionally employed. In this case, an additional \ac{RO} is introduced, where the text is interpreted first in a line-level direction and then typically follows a perpendicular direction to continue reading the subsequent lines, as depicted in Fig. \ref{fig:ro_two_directions}. This structure is commonly known as \textit{paragraphs} and establishes the next complexity level of {HTR}, since a new \ac{RO} is introduced.

\begin{definition}
Paragraph-level HTR is the process of transcribing handwritten text from an input image containing multiple lines of text, where two predetermined \ac{RO} directions must be followed: one for reading individual lines and another for traversing between lines. These directions are commonly perpendicular.
\end{definition}

\begin{figure}[h]
    \centering
    \includegraphics[scale=0.45]{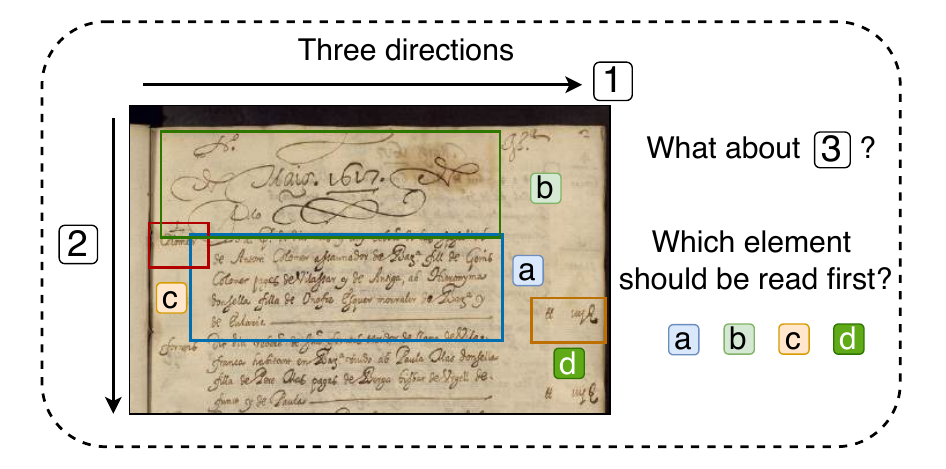}
    \caption{Reading Order (RO) with three directions at the document level. In this case, the third RO is not necessarily consistent or trivial, and each manuscript may define its own rules regarding this aspect.}

    \label{fig:ro_three_directions}
\end{figure}

Ultimately, the highest level of complexity arises when dealing with sources that contain multiple \textit{paragraphs}. The arrangement of these blocks may not be straightforward, as shown in Fig. \ref{fig:ro_three_directions}. In this case, a third \ac{RO} across \emph{paragraphs} is introduced, depending on the layout of each source.

\setlength{\belowdisplayskip}{8pt}

This layout may be either a static geometric arrangement, such as two-column sources, or a dynamic one, where artifacts such as images and tables may be introduced between text blocks. Therefore, we present the final definition.

\begin{definition}
Document-level HTR is the process of transcribing handwritten text from sources that contain multiple paragraphs arranged in an arbitrary, non-trivial \ac{RO}.
\end{definition}

Note that this definition excludes page images where a simple stack of paragraphs is found, which are sometimes understood as ``full-page HTR''~\cite{Coquenet:ICDAR:2021, Singh:ICDAR:2021}. In our view, the transcription challenge remains the same as for a single paragraph, since the \ac{RO} for transitioning between lines also applies to transitioning between paragraphs. Therefore, from a technical perspective, such pages fall within the scope of the definition given for ``paragraph-level HTR''.

Our hierarchical categorization of complexity levels provides a robust framework for understanding and addressing {HTR} challenges. These distinctions are fundamental to the comprehension of {HTR} methods, as different approaches may be better suited to different complexities. Accordingly, we structure the different methodologies based on where the most significant shift in complexity occurs due to the \ac{RO}. Specifically, this refers to the transition from transcription \emph{up to line level} (\textit{words} and \textit{lines}) to \emph{beyond the line level} (\textit{paragraphs} and \textit{documents}).

\subsection{Formulation of HTR}
Let us denote by $\mathcal{X}$ the space of images containing handwritten text. Each element of $\mathcal{X}$ is represented as a three-dimensional tensor in $\mathbb{R}^{H \times W \times C}$, where $H$, $W$, and $C$ represent the height, width, and number of channels of the image, respectively, and their dimensions may vary across instances. Given an input $\mathbf{x} \in \mathcal{X}$, the goal of HTR is to retrieve its corresponding textual representation $\mathbf{y} = (y_1, y_2, \ldots, y_N)$, where $y_i \in \Sigma$ denotes individual \emph{characters} from a predefined \emph{alphabet} $\Sigma$. The alphabet $\Sigma$ may include alphanumeric characters, punctuation symbols, or other special tokens such as spaces or line breaks. Statistically, this retrieval task can be formally defined as:
\setlength{\abovedisplayskip}{6pt}
\setlength{\belowdisplayskip}{6pt}
\begin{equation}
\label{eq:htr-map}
\mathbf{y^*} = \arg\max_{\mathbf{y} \in \Sigma^{*}} \text{P}(\mathbf{y} \mid \mathbf{x}),
\end{equation}

where $\Sigma^{*}$ is the Kleene closure over the alphabet $\Sigma$, representing all possible sequences of characters.

It is known that solving Eq.~\ref{eq:htr-map} exactly is computationally intractable \cite{delahiguera2014most}. Therefore, most HTR methods rely on approximate inference strategies to solve this problem efficiently. These methods differ in how they approximate Eq.~\ref{eq:htr-map} and how they estimate the probabilities involved.

Using Bayes' theorem, Eq.~\ref{eq:htr-map} can be reformulated as:
\setlength{\abovedisplayskip}{6pt}
\setlength{\belowdisplayskip}{6pt}
\begin{equation}
\label{eq:htr-map-bayes}
\text{P}(\mathbf{y} \mid \mathbf{x}) = \frac{\text{P}(\mathbf{x} \mid \mathbf{y}) \, \text{P}(\mathbf{y})}{\text{P}(\mathbf{x})}.
\end{equation}

Since $\text{P}(\mathbf{x})$ is independent of $\mathbf{y}$ during the maximization process, the problem can be equivalently expressed as:

\begin{equation}
\label{eq:htr-bayes-max}
\hat{\mathbf{y}} = \arg\max_{\mathbf{y} \in \Sigma^{*}} \text{P}(\mathbf{x} \mid \mathbf{y}) \, \text{P}(\mathbf{y}),
\end{equation}

where:
\begin{itemize}
    \item $\text{P}(\mathbf{x} \mid \mathbf{y})$ represents the \emph{likelihood}, which models how well the transcription $\mathbf{y}$ explains the input image $\mathbf{x}$. This is implemented using a statistical model, such as a neural network.
    \item $\text{P}(\mathbf{y})$ represents the \emph{prior probability} of the transcription $\mathbf{y}$, often estimated using a language model (LM).
\end{itemize}

The prior probability $\text{P}(\mathbf{y})$ can be modeled using statistical LM such as $n$-grams, where $n$ defines the context size. A LM is formally defined as a function:

\[
p : \Sigma^{*} \to [0, 1], \quad \text{such that} \quad \sum_{\mathbf{y} \in \Sigma^{*}} P(\mathbf{y}) = 1.
\]

LMs provide prior knowledge about the structure of valid sequences and are often integrated into the recognition process through techniques such as re-scoring. Alternatively, in some domains, stricter constraints can be imposed by using a formal language or \emph{lexicon}. Let $\mathbf{L} \subset \Sigma^{*}$ denote the set of valid transcriptions defined by such constraints. In this case, the HTR problem is reformulated as:
\setlength{\abovedisplayskip}{10pt}
\setlength{\belowdisplayskip}{10pt}
\begin{equation}
\label{eq:htr-map-language}
\hat{\mathbf{y}} = \arg\max_{\mathbf{y} \in \mathbf{L}} \text{P}(\mathbf{x} \mid \mathbf{y}) \, \text{P}(\mathbf{y}).
\end{equation}

At a formulation level, the challenge in HTR is consistent across different levels of complexity, but the alphabet $\Sigma$ changes depending on such level: at the \emph{character level}, $\Sigma$ represents the basic charset; at the \emph{word level}, $\Sigma$ includes additional symbols such punctuation marks; at the \emph{line level}, $\Sigma$ incorporates spaces; at the \emph{paragraph level}, $\Sigma$ includes line breaks; and at the \emph{document level}, $\Sigma$ might be further expanded to include paragraph breaks or other structural tokens. However, regardless of the level, the goal of HTR is still to approximate $\text{P}(\mathbf{y} \mid \mathbf{x})$.

% TODO: Comentar aquí (por aclarar) que niveles más bajos siempre tienen un subconjunto del vocabulario del siguiente nivel. Word-level: charset; line-level: charset + space ...

%$\Sigma$ may have a number of special characters belonging to that level, such as space in the case of line level, line break at paragraph and document level, etc. Consequently, the resulting $\Sigma'$ would be the union of the $\Sigma$ with the special characters of the level, defined as $\Sigma' = \Sigma \cup \Xi$.

\subsection{Relation with other fields}
\begin{figure*}
    \centering
    \includegraphics[width=\linewidth]{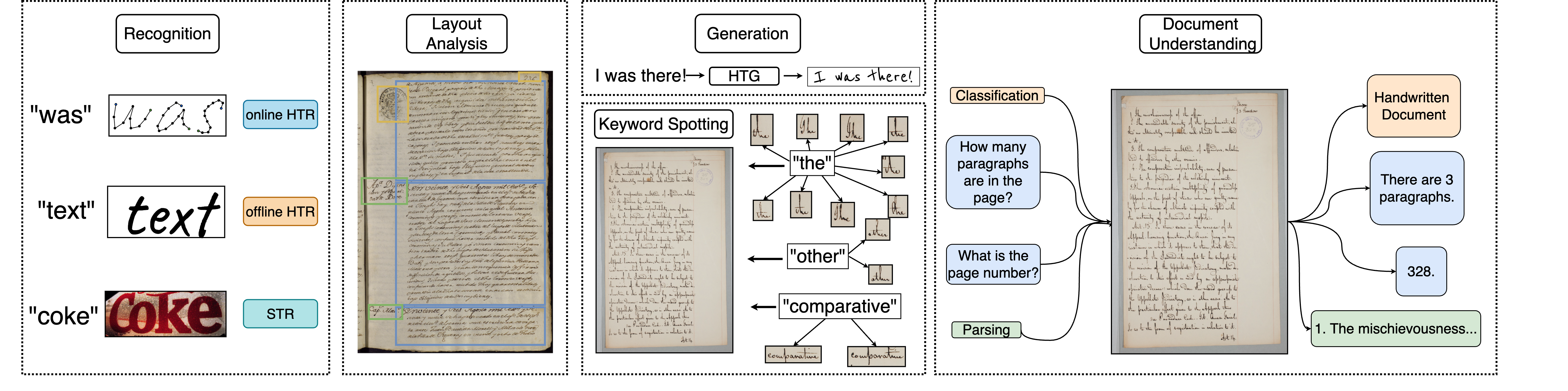}
    \caption{Overview of related fields in Handwritten Text Recognition (HTR). The figure illustrates key areas closely related to HTR, including: (1) Text Recognition (online HTR, offline HTR, and Scene Text Recognition - STR); (2) Layout Analysis (LA) for segmenting and classifying document regions; (3) Handwritten Text Generation (HTG); (4) Keyword Spotting (KWS) for targeted retrieval of specific words; and (5) Document Understanding (DU), encompassing classification, parsing, and semantic analysis of handwritten documents to extract structured information such as paragraph counts, page numbers, and textual content.}
    \label{fig:related_fields}
\end{figure*}

HTR relates to diverse domains, many of which share both challenges and techniques. In this section, we introduce the fields most related to the focus of this survey, placing special emphasis on the differences that make HTR worthy of specific investigation. We illustrate the differences between HTR and the related fields in Fig. \ref{fig:related_fields}.

\subsubsection{Online HTR}
%Online, or pen-based HTR \cite{advances_ghosh_2022, scalable_ingle_2019, online_namboodiri_2004}, involves the recognition of text as it is written using a digital pen or stylus. Unlike ``offline'' HTR, which deals with static images of handwritten text, online HTR leverages dynamic temporal information such as pen tip coordinates, pressure, and tilt. This temporal data provides rich contextual clues about the writing process, including stroke order and direction, which are absent in static images.

%The primary difference between online and offline HTR lies in the nature of the input data. Online HTR can capture the sequence of pen movements, allowing it to utilize dynamic features for higher accuracy. This dynamic information can be rendered into static images, effectively converting online HTR into an offline HTR problem. However, the reverse is not possible. This distinct focus on processing static images and overcoming the challenges associated with them makes offline HTR a specialized field that requires dedicated research and techniques. Additionally, the vast majority of historical documents, forms, and notes that require digitization are in static form, making offline HTR a more widely applicable field of study.

Online, or pen-based HTR \cite{advances_ghosh_2022, scalable_ingle_2019, online_namboodiri_2004}, involves recognizing text as it is written using a digital pen or stylus, leveraging dynamic temporal information such as pen tip coordinates, pressure, and tilt. This data provides contextual clues about stroke order and direction, which are absent in ``offline HTR''. While online HTR can be converted into an offline problem by rendering pen movements into images, the reverse is not possible. Offline HTR requires handling handwriting variability without access to temporal data, making it a distinct challenge with broader applicability, particularly for historical documents and scanned notes.

\subsubsection{Scene Text Recognition}
%Scene Text Recognition (STR) \cite{shi_2015, read_fang_2021, text_chen_2021} focuses on recognizing text embedded in natural scenes captured in photographs or video frames \cite{naiemi2022scene}. STR must contend with diverse and unpredictable environments, including varying text font styles, distortions, and complex backgrounds. These challenges are distinct from those faced in HTR, where the primary concern is the variability in handwriting styles within (relatively) controlled document images.

%While both STR and HTR leverage similar advances in pattern recognition and machine learning, the specific techniques and models developed for STR are tailored to address the complexities of text in natural environments. In contrast, HTR focuses on transcribing handwritten text from static document images, requiring models to handle a wide range of handwriting styles and contextual variations within the text. The controlled nature of document images in HTR and the absence of background noise and environmental factors clearly distinguish it from STR.
Scene Text Recognition (STR) \cite{shi_2015, read_fang_2021, text_chen_2021, naiemi2022scene} focuses on recognizing text in natural scenes, where varying fonts, distortions, and complex backgrounds pose unique challenges. While both leverage similar advances in pattern recognition and computer vision, the primary challenge in HTR lies in handling handwriting variability, while STR must address environmental noise and background interference, making them distinct fields despite their methodological overlaps.

\subsubsection{Keyword Spotting}
Keyword Spotting (KWS) \cite{icfhr2016_pratikakis_2016, deep_wicht_2016, hmm_toselli_2016, novel_frinken_2012} involves identifying and locating specific keywords within a set of document images. This field is particularly relevant for searching and retrieving information from large collections of handwritten documents. 
While KWS and HTR share the goal of identifying handwritten text, they approach the problem from different angles. HTR focuses on the comprehensive transcription of handwritten text into machine-readable format, while KWS is concerned with locating specific words or sentences within the text without necessarily transcribing the entire document.

\subsubsection{Layout Analysis}
Layout Analysis (LA) \cite{docbank_li_2020, doclaynet_pfitzmann_2022, document_binmakhashen_2019} focuses on segmenting and categorizing document components such as text blocks, tables, and images to understand their spatial arrangement. While both LA and HTR deal with document images, they address different aspects of document interpretation. LA focuses on the structural analysis and segmentation of the document, ensuring that various elements are correctly identified and classified. In contrast, HTR specifically concentrates on transcribing the handwritten text within these segmented areas.

\subsubsection{Document Understanding}
Document Understanding (DU) \cite{docformer_appalaraju_2021, unidoc_gu_2021, document_vanlandeghem_2023, ocrfree_kim_2021} encompasses a comprehensive set of tasks aimed at interpreting and extracting meaningful information from documents. This includes text recognition, entity extraction, or document question answering. While HTR provides the fundamental transcription of handwritten text, DU extends beyond this by analyzing relationships between text blocks, extracting key information, and generating structured representations. This requires additional layers of analysis, which are beyond the primary focus of HTR.

\subsubsection{Handwritten Text Generation}
%Handwritten Text Generation (HTG) \cite{scrabblegan_fogel_2020, handwritten_pippi_2023, handwritten_liu_2021, vatr_vanherle_2024} involves the automatic creation of synthetic handwritten text images from digital text input. This field focuses on modeling the variability and fluidity of human handwriting to produce realistic handwritten samples. Techniques such as Generative Adversarial Networks \cite{ganwriting_kang_2020, scrabblegan_fogel_2020, handwritten_liu_2021} and Transformers \cite{handwritten_pippi_2023, vatr_vanherle_2024} are commonly employed to capture the diverse styles and nuances of natural handwriting.

%The primary goal of HTG is to create text that appears as if it were written by a human, which involves challenges such as ensuring that the generated handwriting is realistic and diverse while displaying the expected text. This has applications in personalized handwriting synthesis, CAPTCHA systems, and augmenting datasets for training HTR models. By generating diverse and realistic handwriting samples, it can help improve the robustness and accuracy of HTR systems, especially in scenarios where labeled handwritten data is scarce.

Handwritten Text Generation (HTG) \cite{ganwriting_kang_2020, handwritten_pippi_2023, vatr_vanherle_2024} focuses on synthesizing realistic handwritten text images from digital input, leveraging techniques such as Generative Adversarial Networks \cite{ganwriting_kang_2020} and Transformers \cite{handwritten_pippi_2023, vatr_vanherle_2024}. This has applications in personalized handwriting synthesis, CAPTCHA systems, and augmenting datasets for training HTR models.

% ============================
% === METHODOLOGIES ===
% ============================
\begin{figure*}[ht!]
    \centering
     \includegraphics[width=\linewidth]{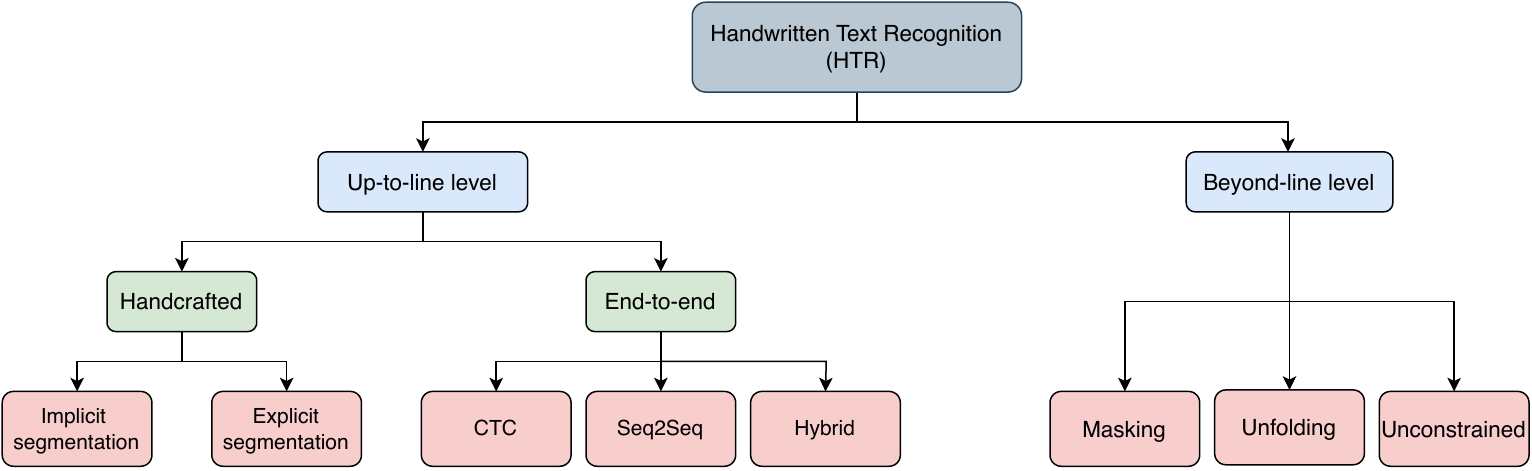}
    \caption{Overview of Handwritten Text Recognition (HTR) methodologies. We divide the approaches into two main groups: up-to-line and beyond line-level. Up-to-line methods are further divided into Handcrafted approaches (explicit and implicit segmentation) and End-to-end techniques (CTC, Seq2Seq, and Hybrid models). Beyond line-level methods include SFR (Segment-Free Recognition) and Segmentation-free techniques, which are further specialized into Masking, Unfolding, and Unconstrained strategies.}
    \label{fig:methodologies-HTR}
\end{figure*}

\section{Methodologies}
\label{sec:methods}
This section outlines the methodologies employed in the field of HTR, categorizing them based on the recognition level: up to line-level and beyond line-level. It examines the historical developments, methodological distinctions, and complementary techniques that enhance performance in the transcription of handwritten text images.
HTR models are a class of probabilistic models that attempt to estimate the most probable transcription $y$ given an image $x$, thus in the form of Eq. \ref{eq:htr-map-language}. 
Conceptually they all address the same task, that of aligning an input image with the most probable output sequence. The differences are encountered in (1) the alphabet ($\Sigma$), which differs depending on the level at which recognition is performed, and (2) how they align the input image with the output sequence. 

We divide this section into two main categories: transcription models up to the line level (word and lines) and models beyond the line level (paragraph and documents).

% ==================================
% ====== UP TO LINE LEVEL ==========
% ==================================

\subsection{Up-to-line level transcription}
\label{sec:up-to-linelevel}
% =====================================
% \begin{itemize}
% \item Up to line level
% \begin{itemize}
% \item Segmentation-based
% \item  Segmentation-free (separar por formas de alinear, que es lo que complica el segmentation-free)
% \begin{itemize}
% \item HMM: generative training, discriminative training
% \item CTC: encoder-decoder (CNN, CRNN, RNN, MDLSTM, Transformer-decoder)
% \item Sequence to sequence: RNN-RNN-Attention, Transformer
% \item Hibridos CTC-seq2seq
% \end{itemize}
% \end{itemize}
% \end{itemize}

%=====================================
% Historical overview (a modo de resumen)
% HMM, CRNN + CTC, Seq2Seq, Transformers, Hybrid. 
% Clasificación en 3 tipos: Encoder-decoder con CTC, no-CTC (Seq2Seq) y arquitecturas híbridas
% \subsection{Handwritten Text Recognition}
% \subsection{Pre-Deep Learning era}

% Aquí hablamos de los métodos hasta up-to-line y 
% Los dividimos acorde a la forma que tienen de alinear (entrenar). 

\begin{figure}
    \centering
    \includegraphics[width=\columnwidth]{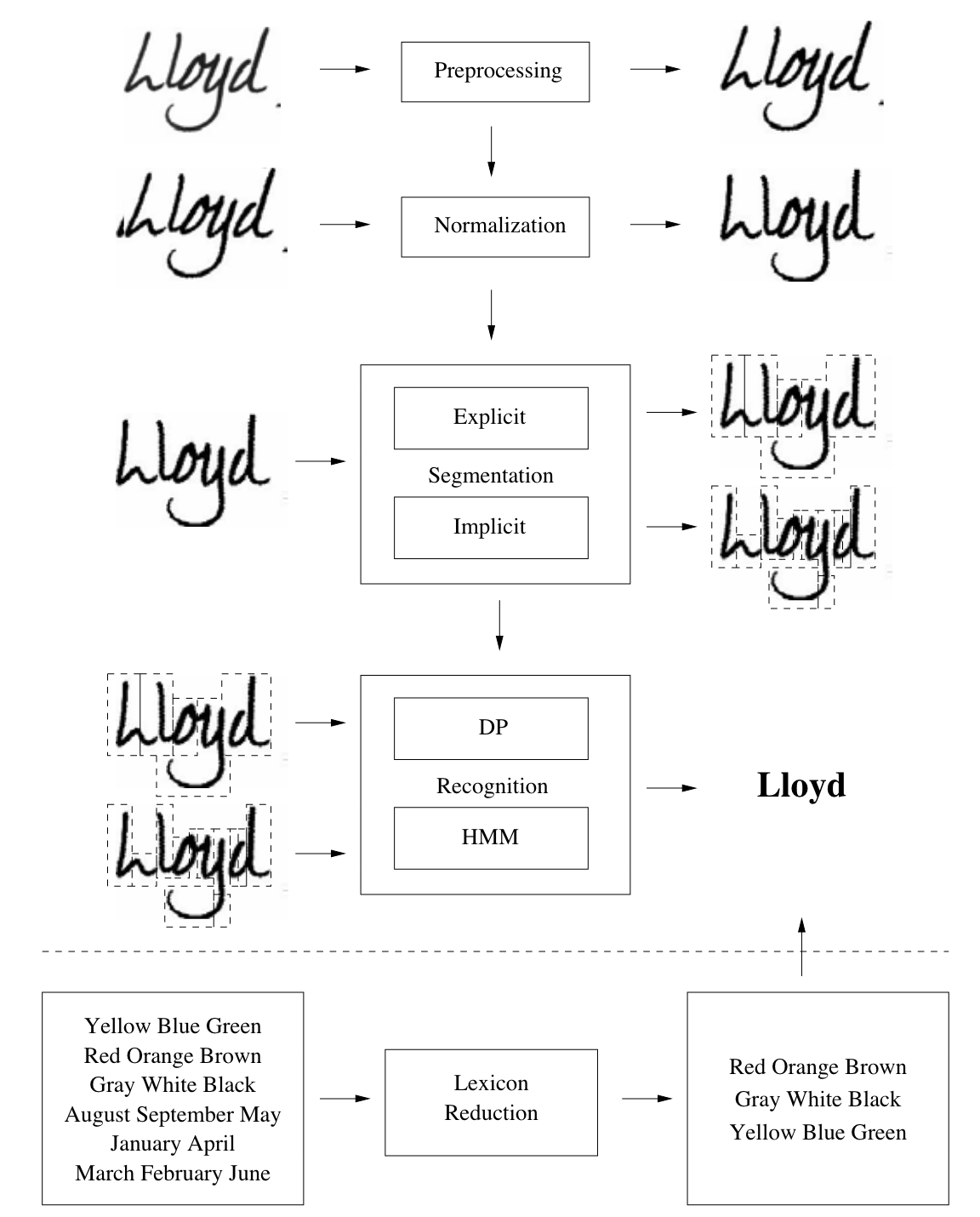}
    \caption{Typical pipeline before the Deep Learning era. Image from \cite{survey_vinciarelli_2002}.}
    \label{fig:classical-pipeline}
\end{figure}

%Still following the original formulation, this level involves transcribing an input image $X$ to an output sequence $y$ composed of an alphabet $\Sigma$. Specifically, 
The challenges \emph{up to line level} involve two categories: (1) word level, in which only characters without spaces are predicted or (2) line level, in which characters are predicted by adding the space character to $\Sigma$, which both share one single RO (one single direction). Before we start to unify the different approaches, we shall make a distinction between the models that go up to the line level. In the early days of HTR, due to the lack of computational and methodological resources, transcription was done in a more manual and laborious pipeline as shown in Fig. \ref{fig:classical-pipeline} and mainly focused on word recognition \cite{multilevel_srihari_1987, recognition_kundu_1989, reading_edelman_1991, handwritten_he_1992, multiclassifier_plessis_1993, improving_boquera_2011, Graves:NIPS:2008, large_koerich_2003, offline_chen_1994, holistic_lavrenko_2004}. % Quitable siguiente párrafo
These steps consisted first of a pre-processing stage where the objective was to remove uninformative elements such as the background, lines or other elements and a posterior binarization to facilitate the extraction of information \cite{survey_vinciarelli_2002, offline_steinherz_1999}. The next step was normalization, which involved “rectifying” the image to keep the image invariant to the writer's style. Typical operations in this case were skew \cite{offline_bozinovic_1995, m_morita_mathematical_1999, myriam_cote_automatic_1998, alessandro_vinciarelli_off-line_2004} and slant correction \cite{offline_bozinovic_1995,alessandro_vinciarelli_off-line_2004,jinhai_cai_off-line_2000,hmmbased_elyacoubi_1999,lexicon_kim_1997, offline_senior_1998, ergina_kavallieratou_slant_2001}, smoothing (denoising) and image scaling. The next step was segmentation (also called fragmentation), in which the information appearing in the word was separated to be further recognized \cite{character_lu_1996, text_louloudis_2009}.

At the methodological level, this does not change in that all still have to estimate Eq. \ref{eq:htr-map-language}, but these factors led to a series of work in which alignment had to occur after a number of manually extracted features. Specifically, these methods differed in whether the alignment was done after isolating the characters from the sequence (explicit segmentation) \cite{survey_casey_1996,character_lu_1996, general_favata_1997,offline_favata_1992,line_burges_1993, handwritten_chen_1993,complement_chen_1994,handwritten_he_1992,offline_chen_1994,oline_senior_2012,writer_gilloux_1994,offline_he_1994,strategies_gilloux_1993,variable_cheii_1993,offline_chen_1992} or whether this separation was done implicitly (implicit segmentation) \cite{offline_saon_1997,
offline_bunke_1994,offline_bunke_1995,reading_edelman_1991,handwritten_caesar_1994,holistic_madhvanath_1996,global_parisse_1996,pruning_madhvanath_1997,modeling_cho_1995,hmm_guillevic_1997,handwritten_mohamed_1996,hidden_gilloux_1994, automatic_paquet_1993, optimised_dewaard_1995,strategies_gilloux_1993,multiclassifier_plessis_1993,strategies_gilloux_1995,writer_gilloux_1994}, so we consider this series of methods as two leaves of the same branch and group them into one: handcrafted methods. Therefore, we will divide up-to-line methodologies according to the ``manual'' learning process: handcrafted (manual) and end-to-end (automatic) approaches.

%%%%%%%%%%%%%%%% AFTER %%%%%%%%%%%%%%%% 
\subsubsection{Handcrafted approaches}
Early methods for HTR relied on handcrafted processes, focusing on segmenting word images into smaller components that could be matched to character models \cite{survey_casey_1996,character_lu_1996}. We summarize this ``classical'' pipeline in Fig. \ref{fig:classical-pipeline}. These segmentation-based approaches required manually designed processes to identify primitive segments, introducing significant challenges due to segmentation errors. Among these approaches, we identify two main categories: (1) explicit segmentation-based methods, which require the individual segmentation of each character, and (2) implicit segmentation-based methods, where the segments do not necessarily correspond to individual characters. Recognition performance depended heavily on accurately matching segments to corresponding characters. According to the segmentation methods, we find: 

% In the first group, an explicit segmentation (each character was segmented) was required using the following paradigms: 
\paragraph{Explicit segmentation} Each character was segmented and used the following methods to perform the recognition.
\begin{itemize}
    \item{Dynamic Programming (DP). DP was a widely adopted technique for matching segments to letters \cite{wordlevel_gader_1999,dynamicprogrammingbased_gader_1996,handwritten_mohamed_1996,lexicon_kim_1997,handwritten_kim_1995,machine_kimura_1997,recognition_kim_1996}. By constructing an optimal path through a cost matrix, DP minimized the mismatch between observed segments and hypothesized letters. The method handled gaps and substitutions effectively but required carefully defined cost functions, often tailored to specific datasets or writing styles.}
    
    \item{Shortest path in graphs. Graph-based methods reformulated the segmentation problem as finding the shortest path in a graph, where nodes represented segment-letter pairs and edges encoded transition costs between them \cite{general_favata_1997,offline_favata_1992,line_burges_1993}. These methods evaluated all possible segment-to-letter matches in a structured way, leveraging graph traversal algorithms to find the least costly alignment. This approach allowed greater flexibility in handling overlapping or ambiguous segments.}
    
    \item{Hidden Markov Models (HMMs). HMMs became a prominent tool in HTR due to their capability to model sequential data with hidden states \cite{handwritten_chen_1993,complement_chen_1994,handwritten_he_1992,offline_chen_1994,oline_senior_2012,writer_gilloux_1994,offline_he_1994,strategies_gilloux_1993,variable_cheii_1993,offline_chen_1992}. Each letter was modeled as a sub-HMM, and these were concatenated to represent entire words. The Viterbi algorithm \cite{viterbi_error_1967} was used to identify the most probable sequence of matches between observed segments and model states. HMMs were particularly effective for handling variability in writing styles and sequential dependencies between letters.}
\end{itemize}

\paragraph{Implicit segmentation}
While early methods relied on explicit segmentation, later approaches explored segmentation-free paradigms \cite{survey_vinciarelli_2002}, where features were learned automatically rather than manually defined. However, these approaches still used handcrafted features, such as:
\begin{itemize}
    \item Low-level features: Basic elements like word contours, strokes, and skeleton approximations of segments \cite{offline_saon_1997,offline_bunke_1994,offline_bunke_1995,reading_edelman_1991,handwritten_caesar_1994,holistic_madhvanath_1996,global_parisse_1996,pruning_madhvanath_1997,modeling_cho_1995,hmm_guillevic_1997}.
    \item Medium-level primitives: Aggregations of low-level features, forming more complex patterns like complete letters or stroke groups \cite{automatic_paquet_1993}.
    \item High-level holistic features: Global characteristics of word images, such as ascenders, descenders, loops, or stroke patterns \cite{optimised_dewaard_1995,strategies_gilloux_1993,multiclassifier_plessis_1993,strategies_gilloux_1995,writer_gilloux_1994}.
\end{itemize}

Despite these advancements, both segmentation-based and segmentation-free methods often relied on alignment techniques like Minimum Edit Distance \cite{global_parisse_1996,holistic_madhvanath_1996,pruning_madhvanath_1997,automatic_paquet_1993,multiclassifier_plessis_1993,optimised_dewaard_1995} or HMMs \cite{hmm_guillevic_1997,offline_saon_1997,offline_bunke_1994,offline_bunke_1995,handwritten_caesar_1994,modeling_cho_1995,hidden_gilloux_1994,writer_gilloux_1994,strategies_gilloux_1993,strategies_gilloux_1995}. These techniques provided a framework for aligning observed features or segments to target words, regardless of the feature abstraction level or segmentation method. Due to the laboriousness of the process, the error-prone series of steps in the classical pipeline and the lack of computational resources, almost all the work described focused on word recognition. 

\begin{figure*}
    \centering
    \includegraphics[width=\linewidth]{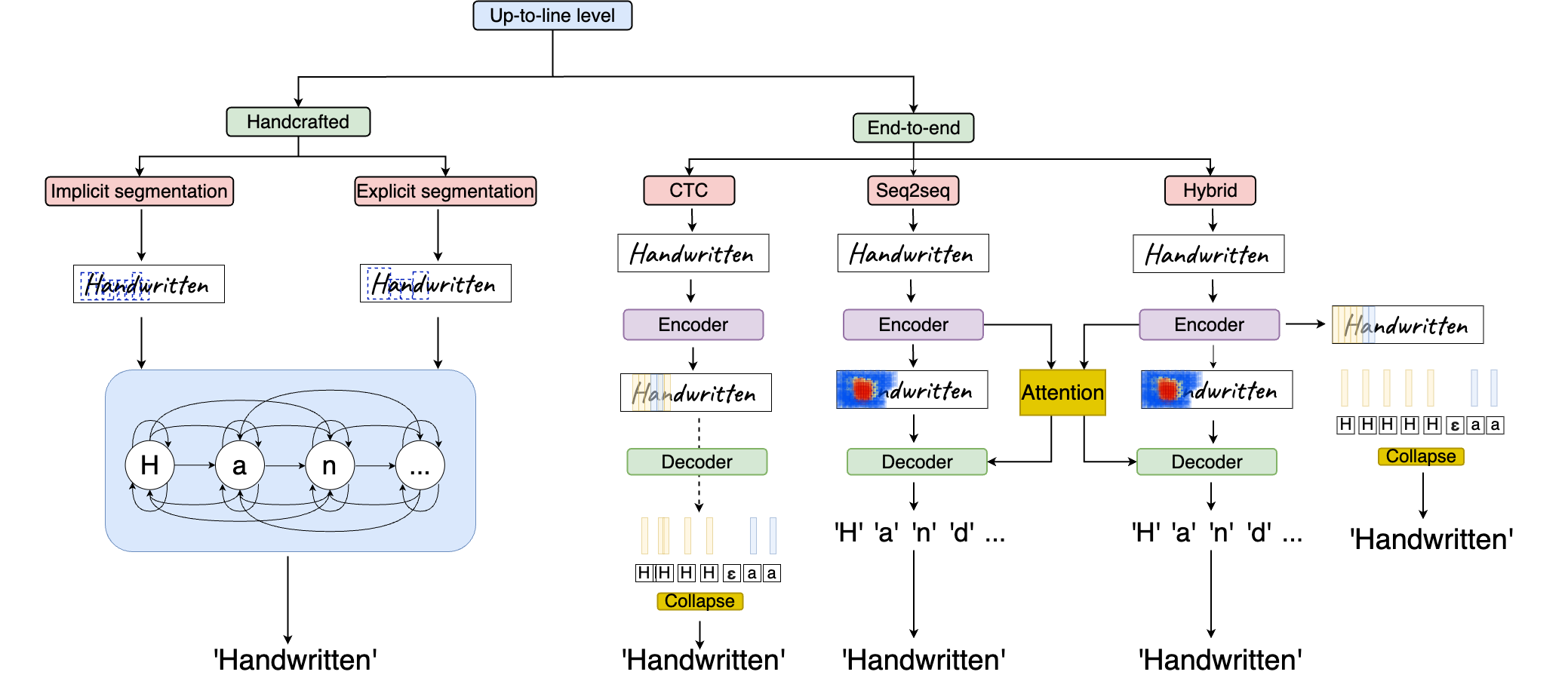}
    % \caption{End-to-end schema of the different approaches.}
    \caption{Taxonomy of HTR approaches up to the line level. Methods are categorized into handcrafted and end-to-end approaches. Handcrafted methods rely on implicit or explicit segmentation, while end-to-end approaches include CT), sequence-to-sequence (Seq2Seq) models with attention, and hybrid architectures. Encoders extract visual features, and decoders reconstruct the textual sequence, with certain models employing attention mechanisms.}
    \label{fig:up-to-line-level}
\end{figure*}

\begin{figure}[ht!]
    \centering
    \includegraphics[width=1.0\linewidth]{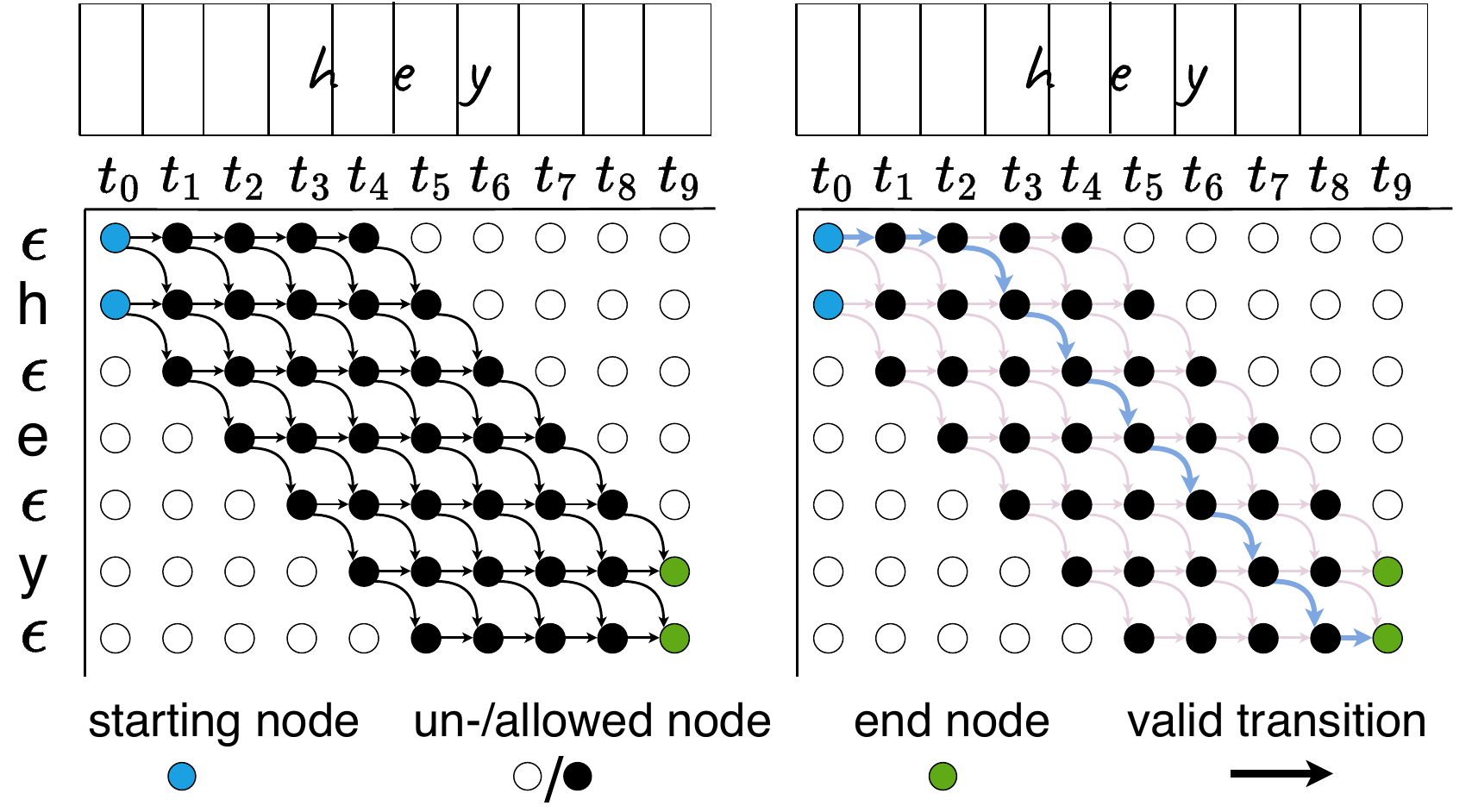}
    \caption{Visualization of the Connectionist Temporal Classification (CTC) training mechanism for HTR. An input image $x$ with the text sequence $y=~'\text{hey}'$ is provided. For simplicity, we exemplify the process with a raw image. The left figure shows the valid transitions between nodes (allowed and end nodes) over time steps ($t_0, t_9$) while processing the handwritten input "hey". Starting nodes, unallowed nodes, allowed nodes, and end nodes are color-coded for clarity. The right figure shows how CTC optimizes the probabilities for the path that maximizes the alignment.}
    \label{fig:ctc_schema}
\end{figure}

\subsubsection{End-to-end approaches}
% Introducir lo que viene siendo un poquito los segmentation-free brother

% Due to the laboriousness of the process, the error-prone series of steps in the classical pipeline and the lack of computational resources, almost all the work described focused on word recognition. (Moved at the end of handcrafted)

% Even those works that process lines directly, take as a previous step word segmentation for a posterior classification of the words. The first work that works directly with lines without word segmentation is found in \cite{marti_handwritten_sentence_2000}, where the proposed system to extract features based on calculated information (such as no. of pixels, momentums, orientations, etc.) using a sliding window, and applied Hidden Markov Models (HMMs) for recognition, integrating a statistical language model to improve performance. 
With the advent of Deep Learning (DL), all features are learned automatically in an end-to-end manner, which is what the field is based on \cite{lecun2015deep}. Under DL-based methods, these differ essentially in how they automatically align the learned features with the output sequence, which we have previously referred to as ``alignment'' i.e. to approximate Eq. \ref{eq:htr-map-language}. Depending on the method employed to align the features with the output sequence, one can find (1) CTC-based methods  \cite{shi_2015, multidimensional_recurrent_puig_2017, Graves2009_Line, improving_boquera_2011, Bluche:NIPS:2016, wigington2018start, offline_sueiras_2018, improving_dutta_2018, boosting_aradillas_2021, Coquenet:ICDAR:2021}, (2) sequence-to-sequence approaches \cite{pay_attention_kang_2022, rescoring_wick_2021, trocr_li_2023, dtrocr_fujitake_2023, rethinking_diaz_2021, attentionhtr_kass_2022} and (3) hybrid techniques \cite{evaluating_michael_2019, light_barrere_2022, training_barrere_2024}. For an outline of the Deep Learning-based methods up-to-line level, see Fig. \ref{fig:up-to-line-level}. Note that, in the HTR literature, we adopt the terminology used in Machine Translation \cite{Bahdanau:ICLR:2015}, referring to the encoder as the architectural component that encodes the maps the input signal into a set of features, while the decoder processes them, generally transcribing in an autoregressive manner. As we will see, some models employ encoder-only architectures, while others integrate both an encoder and a decoder. In the following sections we describe how these different methods align and approximate (\ref{eq:htr-map-language}).
\paragraph{CTC-based methods}
% Given an input signal $x \in \mathcal{X}$, where $\mathcal{X} \in \mathbb{R}^{C \times H \times W}$ (an image in the case of HTR), the objective is to produce a sequence $y$ from a target space $\mathcal{Y} = \Sigma^{*}$, which is defined as all the possible sequences over an alphabet $\Sigma$. These models aim to learn a function $h$ which maps the input space $\mathcal{X}$ to $\mathcal{Y}$ ($h : \mathcal{X} \rightarrow \mathcal{Y}$). 

In order to align the image $x$ with the sequence $y$, CTC introduces the notion of ``valid alignments'' of $y$, which are all the possible character sequences of the alphabet $\Sigma' = \Sigma \cup \{\epsilon\} $ such that collapsing results in the output sequence $y$. This collapse is produced by merging repeated characters and removing the $\epsilon$ (typically named blank) token. Therefore, with CTC, for each labelled sequence $y$ there is a set of $\mathcal{B}^{(y)} \in \mathcal{B}$ valid alignments where $\mathcal{B}$ represents the complete space of alignments. Note that the length of the alignments is given by $\vert x \vert$, which typically corresponds to the number of columns $x_1, x_2, \dots x_T$ of the image, typically referred to as frames.\footnote{In the case of images, each frame is typically associated with a column of the image. This convention originates from the field of ASR, where the continuous nature of audio signals led to their partitioning into overlapping segments known as frames.} In this section, the methodologies explored assume a frame-wise reading. This means that all the frames contained in the feature map represent exclusively one character. This allows up-to-line methods to only read in a single RO direction. To reach this assumption, all methods resort to an \emph{image collapse} step, where data is transformed from an image space to a sequence one. This is defined as a function $r: \mathbb{R}^{w,h,c} \rightarrow \mathbb{R}^{l,z}$, where $h$ is the height of the feature map, $w$ its width, $c$ the number of channels, $l$ is the output sequence length and $z$ the number of features. This function reshapes the input feature map to merge their features along the horizontal axis. Therefore, they address the following general formula:
\setlength{\abovedisplayskip}{8pt}
\setlength{\belowdisplayskip}{8pt}
\begin{equation}
\label{eq:htr-utlt}
\hat{\mathbf{y}} = \arg\max_{\mathbf{y} \in \mathbf{L}} \text{P}(\mathbf{y} \mid r(f(\mathbf{x})))
\end{equation}

where $f(\cdot)$ is the feature extraction process and $r(\cdot)$ the \textit{image collapse} function. For simplicity, from now on we will assume that each frame $x_t$ is the collapsed image into frames.  

The approach for CTC then consists of dividing the input sequence into $T$ frames and predicting, for each frame $t 
\in T$, the probability of each character of the target alphabet. Therefore, the alignment occurs monotonically from left to right, mapping one frame to a single character.

This process is illustrated in the Fig. \ref{fig:ctc_schema}, where the valid alignments correspond to those paths that pass through valid nodes via the transitions established in the diagram. The transition probabilities can be used to infer the actual output in a greedy form by taking the most likely character at each time step $t$ or by using beam search (see Section \ref{subsec:complementary_techniques}).
%, the former being more common
Formally, for a given pair of $(x, y)$ CTC \cite{connectionist_graves_2006} computes the conditional probability 
\begin{equation}
P(y \vert x) = \sum_{\mathcal{B}^{(y)} \in \mathcal{B}}   \prod^{T}_{t=1}P(a_t \vert x_t)
\label{eq:ctc1}
\end{equation}

where $P(a_t \vert x)$ is the probability for a single alignment step-by-step (with $T$ given by the length of the input to align), marginalized over the set $\mathcal{B}^{(y)}$ of valid alignments for that sequence, where $\vert \mathcal{B}^{(y)} \vert < T$. This probability is maximized throughout the log-loss $\sum_{(x,y) \in \mathcal{D}_{train}} -\text{log} P(y \vert x)$, where $\mathcal{D}_{train}$ is the training dataset, often consisting in pairs of images and transcriptions. For inference, it is selected the path that maximizes the alignment $B$:
\begin{equation}
B^{*} = \arg\max_{B}  \prod^{T}_{t=1}P(a_t \vert x_t)
\label{eq:ctc2}
\end{equation}

Typically, the CTC-based architectures used to estimate probabilities $P(a_t \vert x)$ in the HTR field have been LSTMs \cite{Hochreiter-LSTM} and its bidirectional variant (Bi-LSTM) \cite{Bi-LSTMGRAVES2005602}. Specifically, in early work using LSTMs for HTR, the use of Multidimensional LSTMs (MDLSTM) was applied directly on HTR images to recognize words without using any prior feature extractor \cite{Graves2009_Line}. In later works \cite{Pham_dropout_2014, Bluche:ICDAR:2017, gated_bluche_2017}, these MDLSTMs were applied on low-level features obtained by a convolutional feature extractor. The use of different MDLSTM layers with different heights allowed converting 2D signals to 1D allowing to apply CTC to this 1D output. The first attempt to recognize unconstrained lines without the pre-segmentation of words was in \cite{Graves2009_Line}, where the authors used a Bi-LSTM and CTC directly to recognize off-line English texts. Following \cite{shi_2015}, the work of Puigcerver \cite{multidimensional_recurrent_puig_2017}  proposed to make this conversion from a 2D to 1D signal by collapsing the features in height by passing them to image channels to directly recognize sentences without the use of MDLSTMs. This allowed BLSTM-type networks to be applied directly to 1D signals, significantly reducing the computational cost and obtaining better results in terms of transcription. This has been the dominant pipeline for CTC-based models from then \cite{multidimensional_recurrent_puig_2017, Coquenet:ICDAR:2021, kang_convolve_2018, improving_dutta_2018, candidate_kang_2019, evaluating_michael_2019}. 

Moving away from the use of recurrent networks as part of the encoder, there have been new trends to replace this part with Transformer encoder type attention based models as in \cite{porwal_self-attention_2022, light_barrere_2022, training_barrere_2024} while still aligning with the CTC objective. Apart from this substitution, there is also literature using fully convolutional approaches to extract features and align directly on these with CTC as in \cite{Yousef:CVPR:2020, Coquenet:ICDAR:2021}.

\paragraph{Sequence-to-sequence}
\label{sec:seq2seq}
Sequence-to-sequence models typically follow a decoupled architecture consisting of an encoder $g$, which reads the input signal and produces a representation $h = g(x)$, and a decoder $f$, which receives this representation and decodes the output as $y=f(g(x))$. In this case, unlike with CTC, the decoder directly outputs the sequence token by token. Generally, sequence-to-sequence models estimate the probability $P(x \vert y)$ as:

\setlength{\abovedisplayskip}{8pt}
\setlength{\belowdisplayskip}{8pt}
\begin{equation}
P(y \vert x) = \prod^{T}_{t=1} P(y_t \vert y_1, y_2 \dots y_{t-1}, x)
\label{eq:seq2seq1}
\end{equation}

achieved by the chain rule of probability. More specifically, given the encoder $g$, the decoder $f$ estimates the probability given by $P(y_t \vert y_1, y_2 \dots y_{t-1}, h)$, where $h = g(x)$. Typically, $g$ is based on CNN \cite{pay_attention_kang_2022, candidate_kang_2019} or in combination with recurrence \cite{kang_convolve_2018, offline_sueiras_2018, chowdhury2018efficient} or a self-attention mechanism \cite{pay_attention_kang_2022, trocr_li_2023, transformer_wick_2021}. The learning is achieved similarly to CTC with the log-loss 
\(\sum_{(x,y) \in \mathcal{D}_{train}} -\text{log} P(y \vert x)\), 
which is typically implemented using the Cross-Entropy (CE) loss. Mathematically, CE is defined as:
\[
\mathcal{L_{\text{ce}}} = -\sum_{t=1}^{T} \sum_{k=1}^{K} y_{t,k} \log (\hat{y}_{t,k}),
\]
where \(y_{t,k}\) is a one-hot encoded vector representing the true label at timestep \(t\) for class \(k\), 
\(\hat{y}_{t,k}\) is the predicted probability for the same class, \(T\) is the number of timesteps, and \(K\) is the number of classes. CE measures the difference between the predicted probability distribution and the true labels, encouraging the model to assign higher probabilities to the correct outputs. In this approach, the alignment is established from all the frames/columns to a single token, unlike in CTC, where the alignment strictly follows a one-frame-per-token mapping.

In this case, the conditional probability is maximized throughout training using the Teacher Forcing technique \cite{teacher_forcing_williams_1989}, where the correct token \(y_t\) is fed to the model at the \(t\)-th timestep. 

% The learning is achieved as in CTC with the log-loss $\sum_{(x,y) \in \mathcal{D}_{train}} -\text{log} P(y \vert x)$, but in this case this conditional probability is maximized throughout training with the Teacher Forcing technique \cite{teacher_forcing_williams_1989}, where the correct token $y_t$ is fed to the model at the $t$ timestep. 

Generally, an attention mechanism is applied at the decoder side to gather contextual information from $h$ and learn where to find the relevant content from the encoder. Depending on the manner in which they attend to $h$, attention can be mainly divided into two types: (1) content-based attention \cite{Bahdanau:ICLR:2015}, where the network is allowed to freely search for patterns to recognize the character to be predicted and (2) location-based attention \cite{chowdhury2018efficient}, where the network also incorporates information about the locations previously consulted.

In the context of HTR, the first models using sequence-to-sequence models with only minor variation in the attention mechanism can be found in \cite{offline_sueiras_2018,characterbased_poulos_2017,candidate_kang_2019}. 
% In \cite{characterbased_poulos_2017}, they compare softmax and sigmoid content-based attention mechanisms vs. no attention for recognizing lines. The work of \cite{chowdhury2018efficient} also uses the same softmaxed-attention, also incorporating the Loung \cite{luong-etal-2015-effective} mechanism and training with the Focal Loss \cite{focal_loss_lin_2017} as well as layer/batch normalization layers to recognize lines. Also, this work incorporates the technique of beam search \cite{graves_2012_seq}. Meanwhile in \cite{offline_sueiras_2018}, the authors divide the image into windows, using these to obtain the encoder features (independently for each patch) and subsequently apply the standard attention mechanism as in the other works, but applied to recognize handwritten words. Finally, the work in \cite{candidate_kang_2019} also studies the attentions evaluated in \cite{characterbased_poulos_2017}, incorporating the label smoothing \cite{inception_label_smooth_szegedy_2016} mechanism, also to recognize words. As a further and comprehensive review of all attention approaches in sequence-to-sequence models, \cite{evaluating_michael_2019} explores the different attention mechanisms and proposes the use of hybrid training: CTC in the encoder and log-loss for the decoder.

Beyond the ``classic'' sequence-to-sequence models, the architecture that has been dominating the sequence-to-sequence arena is the Transformer architecture \cite{TransformerVaswani}. While remaining conceptually the same as the classical models discussed above, the introduction of the self-attention layer as well as the cross-attention mechanisms turned this architecture into an improved, parallelizable and scalable version of the classical models using attention. However, the need for pre-training or synthetic data is crucial. Transformers were first used in the context of HTR in \cite{pay_attention_kang_2022}, in which the authors propose to train a large Transformer model with the features obtained by a ResNet as a convolutional extractor. In \cite{light_barrere_2022}, they demonstrate that a considerable reduction of parameters both in convolutional extractor and Transformer architecture trained with CE performs equally or slightly better than a large model as in \cite{pay_attention_kang_2022}.

In \cite{transformer_wick_2021}, the authors train two Transformers, one from left-to-right and a right-to-left combining the predictions in a post-decoding process. 
Finally, relying heavily on synthetic data and pre-trained models, TrOCR \cite{trocr_li_2023} uses a fully Transformer-based architecture originally designed for general text recognition. With the same philosophy, DTrOCR \cite{dtrocr_fujitake_2023} removes the encoder part and achieves higher recognition accuracy using nearly 2B of synthetic images for pre-training. The use of synthetic data generation will be studied in further sections. 

\paragraph{Hybrid architectures}
% So far, we have examined various approaches for aligning input images with output sequences, focusing on methods that use CTC alignment or sequence-to-sequence that minimize the CE loss. However, t
There have been a number of works that have opted to use a hybrid approach \cite{evaluating_michael_2019, light_barrere_2022, training_barrere_2024} by combining CTC with the sequence-to-sequence. Building on the work of \cite{kim_hybrid_speech_2017}, who introduced a hybrid method for speech recognition by combining CTC for the encoder and CE for the decoder, this approach was first adapted for HTR in \cite{evaluating_michael_2019}. In their study, the authors explore the integration of visual features learned by the encoder using CTC, with the output sequence predicted by the decoder utilizing these learned features. This is achieved by employing a hybrid loss function that merges both CTC and CE losses as follows:

\setlength{\abovedisplayskip}{12pt}
\setlength{\belowdisplayskip}{12pt}
\begin{equation}
\mathcal{L} = \lambda\mathcal{L_{\text{ctc}}} + (1 - \lambda)\mathcal{L_{\text{ce}}}, \lambda \in  [0, 1]
\label{eq:hybrid-htr}
\end{equation}

where $\lambda$ is a weighting parameter that controls the relative contribution of each loss to the overall optimization process and is typically set to 0.5. This combined loss approach has shown to be advantageous in HTR contexts, as demonstrated in \cite{evaluating_michael_2019}. In this latter approach, images are aligned with sequences using both CTC and sequence-to-sequence alignment mechanisms. In this setting, the encoder is typically unused for inference and the output is predicted as in a sequence-to-sequence approach. Recently, the concept of transcribing as a multitask learning problem has gained traction, particularly with architectures that incorporate lightweight CNNs and Transformers. This trend is exemplified in the works \cite{light_barrere_2022, training_barrere_2024}, which have explored these architectural combinations in addition to synthetic data generation for pre-training to further enhance transcription performance.

% ==================================
% ====== BEYOND LINE LEVEL =========
% ==================================

% \subsection{Beyond Line-Level Transcription}
\subsection{Beyond-line level transcription}
\label{sec:bll_description}

The maturation of up-to-line level transcription techniques in \ac{HTR} has catalyzed a significant paradigm shift toward more ambitious challenges: the direct transcription of complete paragraphs and documents without requiring preliminary segmentation steps for isolating lines. This evolution represents not merely an incremental advancement but a fundamental reconceptualization of the transcription process.

A critical technical barrier in this transition stems from the vertical collapse function inherent to up-to-line level methods. While this approach effectively processes single lines by assuming that vertical features correspond to individual characters, it becomes problematic when scaling to multiple line scenarios. In paragraph or full-page transcription tasks, individual vertical slices of the image frequently contain several characters from different lines that must be independently recognized. The collapse function, by compressing these multiple characters into a single feature representation, fundamentally compromises the ability of the model to maintain proper \ac{RO} and accurately distinguish between characters from different lines. Modern beyond-line level approaches specifically address this limitation through methodological adaptations and extensions. Note that the foundation of these methods is still the one that formulates the HTR field. Beyond line-level transcription methods mainly seek to address the $r(\cdot)$ function to retrieve a sequence-like data structure that represents the whole content of the page, aligned with ground truth.

Given the emerging nature of this paradigm and its profound implications for the field, we perform an individual analysis of current methodological approaches and their technical foundations.

\begin{figure*}
    \centering
    \includegraphics[width=1.0\linewidth]{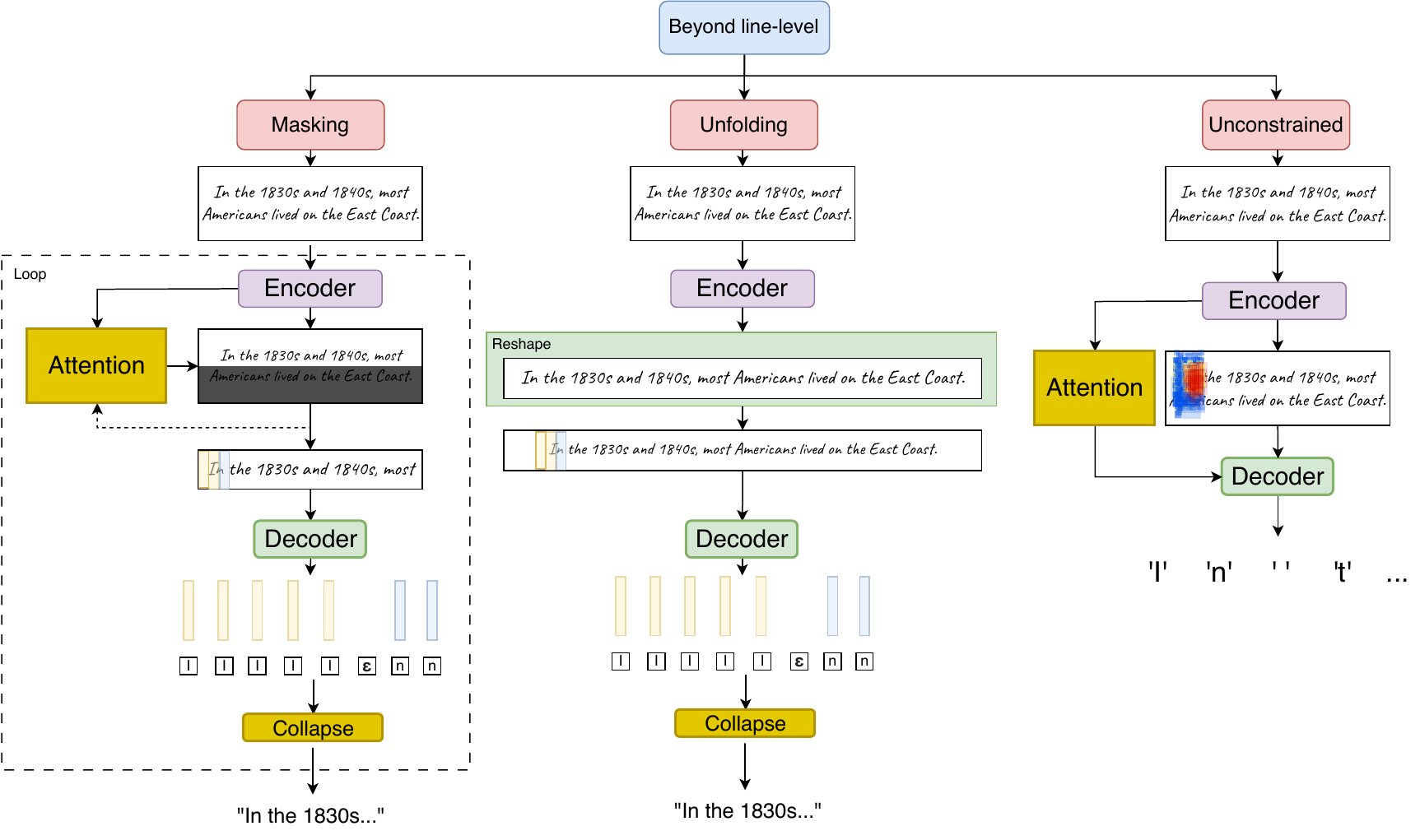}
    \caption{Taxonomy of \ac{HTR} approaches beyond line level. Methods are categorized into masking, unfolding, and unconstrained.}
    \label{fig:beyond line-level}
\end{figure*}

Hereby, we find three different approaches: \textit{attention masking} and \textit{line unfolding}, where methods downscale the complexity of the problem into an up-to-line level challenge, either through attention-based segmentation or reshape functions, and (ii) \textit{unconstrained} methods, where the already established sequence-to-sequence formulation from up-to-line level is adapted to learn more complex inputs. Figure \ref{fig:beyond line-level} depicts a summary of the reviewed approaches.

\subsubsection{Attention masking}
The earliest publications beyond line-level applications found in the literature are those that apply attention matrices masking as the transcription method. These methods implement $r(\cdot)$ as $r(x) = \alpha \cdot x$, where $\alpha$ is a learned mask that is applied to the input feature map. These methods are applied ``autoregressively'', as the mask has to be recalculated to extract all the lines from the input document, simplifying the multi-line complexity of the problem into a sequential line-by-line transcription. The presented methods in this particular approach propose different ways of computing $\alpha$.

The first publications referred to these methods are the \ac{JLSAT}~\cite{Bluche:NIPS:2016} and the \ac{SAAR}~\cite{Bluche:ICDAR:2017} models. Both methods are based on a combination of Convolutional Neural Networks and \ac{MDLSTM}~\cite{Graves:NIPS:2008}. The first model, the \ac{SAAR}, is a sequence-to-sequence approach where attention and inference are computed at the character level.  The second one, \ac{JLSAT}, is an optimization of the first approach that computes attention at the line level and training with the CTC loss function.

Although these approaches provided groundbreaking results---since they were the first segmentation-free models published---they were based in \ac{MDLSTM} networks, which are inefficient in both training and inference. As a result, these approaches have been left behind over time, thanks to the income of Transformer-based autoregressive architectures.\footnote{These early models were published around the same time as the Transformer, in 2017 (see Fig. \ref{fig:timeline-HTR})} Despite this, the philosophy from these papers still prevails both in unconstrained methods, described in Section \ref{subsec:unconstrained}, and in the \ac{VAN}~\cite{Coquenet:TPAMI:2023b}. This network replaces the slow-to-train MDLSTM blocks with a convolutional network and a line-wise hybrid attention. The \ac{VAN} specifically finds, approximately, the rows of a document that, at least, contain a character to transcribe. This method, although it is more efficient both in training and inference, is limited to transcribe only text paragraphs, as the rows are approximated in the vertical axis, assuming therefore that a row contains the full width of the image. 

\subsubsection{Line unfolding}

In parallel to the development of masking methods, the unfolding systems started being developed. This approach works under a single-step end-to-end philosophy, as state-of-the-art line transcription does. In this case, the  $r(\cdot)$ function is approached through a reshape operation that substitutes the vertical collapse between the feature extraction and the decoding stages. %An example of how these methods are generally implemented is depicted in Figure \ref{fig:line_unfolding_gs}.

%\begin{figure}
%    \centering
%    \includegraphics[width=\columnwidth]{figures/line_unfolding.pdf}
%    \caption{Conceptual design of line unfolding models}
%    \label{fig:line_unfolding_gs}
%\end{figure}

The first publication that addresses this formulation is the Origaminet network~\cite{Yousef:CVPR:2020}. The unfolding method is driven by a network composed of convolutional networks and upsampling operations. Authors assume that the produced line is long enough to contain the input document. Hereby, the network learns how to place its elements into the resulting feature map that is read in the correct RO, driven by the ground truth. This operation is done along the horizontal axis, forcing the network to implicitly learn to read from top to bottom and from left to right. Authors claim, with empirical evidence, that the model is able to transcribe full-page documents with non-trivial ROs. Although this network produced substantial results, it relies on user-defined hyperparameters, such as the line-like structure length and a fixed-size image, suggesting that the model performance is corpus-dependent.

The \ac{SPAN} network came briefly after Origaminet, and it is proposed as an unconstrained upgrade of this network~\cite{Coquenet:ICDAR:2021}. Instead of relying on fixed-sized parameters, authors rely on an automatic unfolding operation that concatenates the rows from the feature map produced by the encoder. The concatenation step is performed from top to bottom. This operation is computed after the decision layer.\footnote{The layer that maps the feature embeddings into the output vocabulary of the network.} This is a relevant feature of the \ac{SPAN} model, as authors stress that, by implementing it this way, backpropagation is always applied in a two-dimensional context, which lets the model distinguish better the location of the different lines of the text. By sticking always to the features of the image, the \ac{SPAN} method is independent of the image size, in contrast to the fixed-size images required by the Origaminet.

Despite setting a more generic unfolding method, SPAN is also limited. Specifically, the automatic unfolding operation only works on text paragraphs. When inputting more complex ROs, the network may be unable to transcribe them, as it is very difficult to find a correct alignment between the image and the ground truth. In the work of Parres et. al~\cite{Parres:ICDAR:2024}, they sort out the limitation by implementing a Swin Transformer with the pre-trained weights from DONUT to the SPAN concept, leveraging feature extraction to self-attention instead of convolutions. Although this is not deeply studied in the cited work, the authors empirically prove that this architecture is able to avoid this problem in some corpora.

\subsubsection{Unconstrained}
\label{subsec:unconstrained}

Unconstrained models approximate Eq. \ref{eq:seq2seq1}, but with multi-line input documents. These methods, as they are based on conditioned LMs, are free of any direct correlation between the input features and the ground truth, as in Machine Translation. Therefore, unconstrained models are able to learn an arbitrary RO. These methods arose after the conception of the Transformer architecture, as LSTM-based models did not prove fruitful on the task.

The first published work about unconstrained methods is the \ac{FPHR}~\cite{Singh:ICDAR:2021}. This work proposed two key elements for these methods: (i) the usage of 2D Positional Encoding to hint the language model where each part of the image is bearing in mind spatiality and (ii) extensive synthetic pre-training to be able to learn arbitrary ROs. Nearby this publication, the TBPHTR \cite{Rouhou:PRL:2022} and the \ac{DAN}~\cite{Coquenet:TPAMI:2023} were released, which holds on the same architecture concept, but with lighter implementations. The latter model is the most popular network in the community, as it is open-source, it is able to give structural details about the input document and it was proven state-of-the-art in three publicly-available datasets. Later, the MSdocTr-Lite was proposed with some optimizations of the \ac{DAN} network~\cite{Dhiaf:PRL:2023}. In general, all of these systems follow the same implementation, where a generic CNN is placed as a feature extractor and a Transformer decoder acts as the language model that correlates the previous inputs with the feature map outputted by the encoder.

% \subsubsection{Teaching unconstrained methods to read}
The most relevant differences lay in how these models are trained to handle multi-line documents. Transformers are models with a low inductive bias, meaning that they can learn any task if given enough data. However, most of the data sources HTR provides are scarce in quantity. As mentioned before, the \ac{FPHR} established that a way of sorting this is by pre-training the model with synthetic data. This work, specifically, implements a brute-force training strategy, where synthetic documents are rendered with handwritten fonts, given WikiText content, or through the IAM database cropped words.

In the TBPHTR~\cite{Rouhou:PRL:2022}, authors resort to first pre-train on text-line images from the target dataset and, then, fine-tune the model to transcribe multi-line sources. This has a potential limitation, since it requires that the dataset labeling provides line-level information. However, it establishes an interesting takeaway: it is convenient to first pre-train on text lines, the simplest \ac{RO} category, and then fine-tune into more complex structures seems to solve the task and leverage the data requirements for the model.

The \ac{DAN} \cite{Coquenet:TPAMI:2023} expands further this idea by implementing a multi-stage Curriculum Learning (CL) training. First, the encoder network is trained to transcribe synthetic text lines with the CTC loss function. Then, the whole architecture is trained with synthetic documents of incrementally increasing text lines. As line quantity progresses, the synthetic generator creates increasingly complex documents. Once thirty lines are reached, a final CL is performed, where the model is fed with both synthetic and real samples. The quantity of synthetic and real data fed to the model is controlled by a linearly-scheduled probability.

The MSdocTr-Lite \cite{Dhiaf:PRL:2023} simplifies the method proposed by the \ac{DAN}, by first pre-training through text lines and, then, implementing a CL-based on text blocks complexity, which can be done either with a synthetic dataset or a layout-labeled real-world corpus.

\subsection{Complementary techniques}
\label{subsec:complementary_techniques}
In the evolving field of HTR, significant advancements in neural network architectures have been complemented by sophisticated techniques that enhance recognition accuracy and usability. Among these, decoding strategies are crucial, as they bridge the gap between the probabilistic outputs of neural networks and coherent text sequences. Decoding involves selecting the most plausible character sequences from the network's output probability distributions, a task made challenging by the variability in handwriting and the ambiguity in certain characters. Two prominent decoding methods are \textit{greedy decoding} and \textit{beam search decoding}, with the latter often augmented by lexicon-based constraints for improved performance.

\subsubsection{Greedy decoding}
Greedy decoding is a straightforward method for converting a neural network's output into a character sequence. At each timestep \( t \), the method selects the character \( c_t \) with the highest probability, given the input \( x \). This approach maximizes the likelihood of each individual character independently:

\[
c_t = \arg\max_{c} P(c \mid x),
\]

where \( P(c \mid x) \) represents the probability of character \( c \) at timestep \( t \). While greedy decoding is computationally efficient, it fails to consider dependencies between characters across timesteps. Consequently, it is prone to errors when local decisions lead to globally incoherent sequences. For example, ambiguities in handwriting might result in suboptimal predictions if the most probable character at a single timestep does not align with the overall context of the sequence. There are several work that uses greedy decoding \cite{pay_attention_kang_2022, trocr_li_2023, dtrocr_fujitake_2023, kang_convolve_2018, improving_dutta_2018}

\subsubsection{Beam Search decoding}
\begin{figure}[ht!]
    \centering
    \includegraphics[scale=0.24]{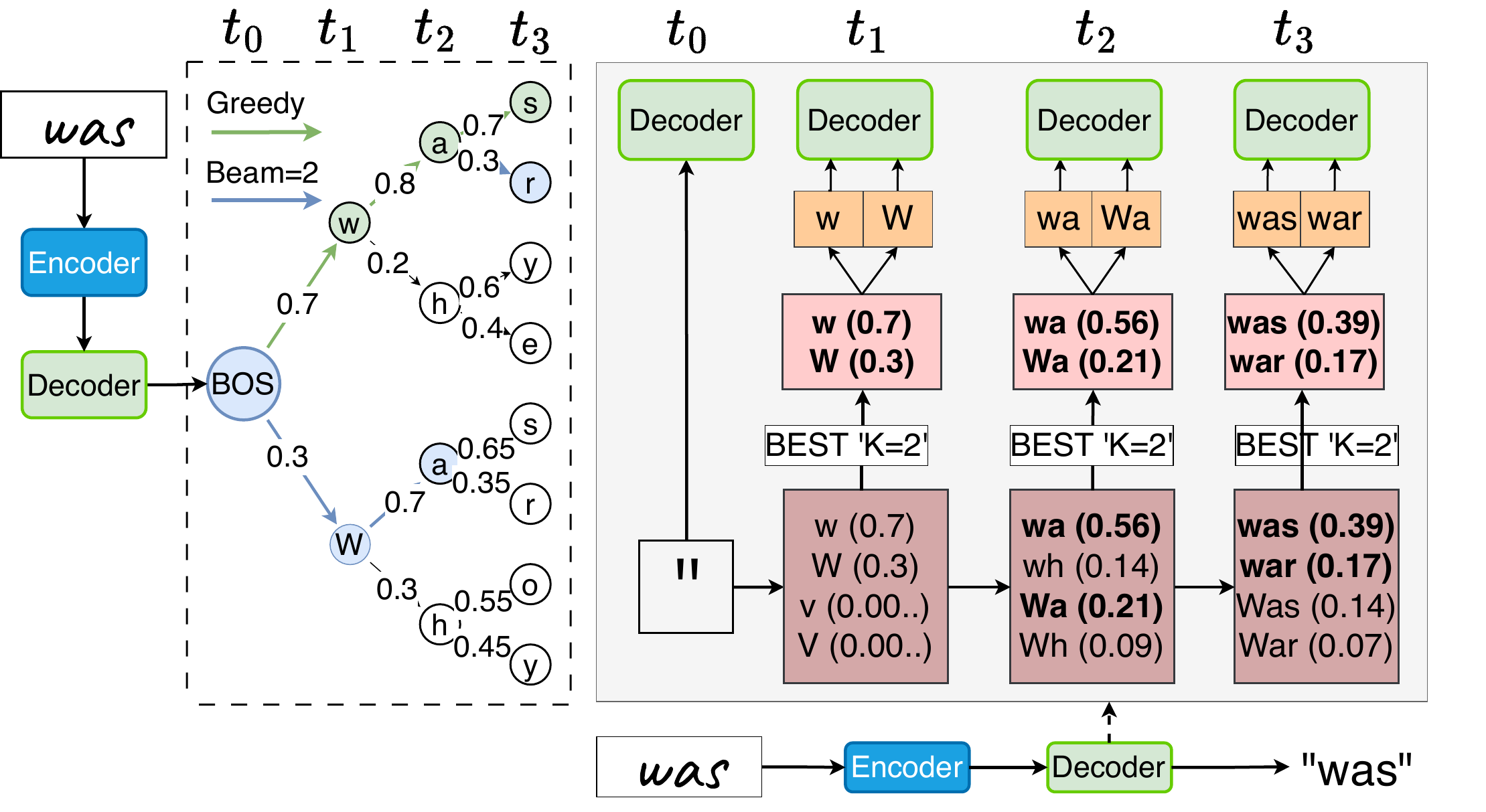}
    % \caption{Illustration of the beam search algorithm with a beam width of \( k = 2 \). The diagram compares greedy decoding, which selects the highest-probability character at each timestep (resulting in the sequence "was"), with beam search, which retains the top-\( k \) most probable sequences throughout the decoding process. At each timestep, beam search evaluates all possible extensions of current hypotheses, ranks them by cumulative probability, and retains the best \( k \) candidates. This approach allows the decoder to recover from local ambiguities and produce contextually coherent predictions. The figure demonstrates the iterative steps of hypothesis selection and pruning and the final sequence ("was") as the highest-scoring output.}
    \caption{Illustration of beam search with beam width \( k = 2 \), compared to greedy decoding. Greedy decoding selects the highest-probability character at each step ("was"), while beam search retains the top-\( k \) most probable sequences, evaluating and pruning hypotheses iteratively. This allows recovery from local ambiguities, yielding the most probable output ("was").}
    \label{fig:beam_search}
\end{figure}

Beam search improves upon greedy decoding by maintaining a fixed number of hypotheses, known as the "beam width" at each timestep. Unlike greedy decoding, which commits to a single character at each step, beam search explores multiple potential sequences in parallel, ranking them by their cumulative probabilities. The probability of a sequence \( s = (c_1, c_2, \dots, c_t) \) is computed as:

\[
P(s \mid x) = \prod_{t=1}^{T} P(c_t \mid x),
\]

where \( T \) is the total number of timesteps. At each time step, the algorithm evaluates all possible extensions of the current hypotheses and retains the top-\( k \) sequences with the highest cumulative scores. This iterative process allows beam search to account for context and sequence-level dependencies, significantly improving transcription accuracy in scenarios with ambiguous handwriting. An example of this process is illustrated in Fig. \ref{fig:beam_search}. Works that use beam search in HTR can be found in \cite{chowdhury2018efficient, evaluating_michael_2019}.

% ============================
% === BENCHMARKING HTR ===
% ============================
\section{Benchmarking}
\label{sec:benchmark}

\subsection{Datasets}
In this section, we provide an overview of the most commonly used Latin-script datasets in HTR literature,\footnote{This overview is not intended to be exhaustive. Instead, we focus on the datasets most frequently used in the HTR literature due to their significant impact, availability, and relevance for advancing the field.} highlighting their key features and contributions to the field. The field of HTR is rich with a variety of datasets although scarce in volume, with each dataset catering to different aspects of the recognition process. These datasets vary significantly in terms of language, historical period, number of writers, and the levels at which they provide annotations (word-level, line-level, or page-level). This diversity is critical for developing robust HTR systems capable of handling various handwriting styles and document structures. Below, we provide an in-depth analysis on some of the prominent datasets used in HTR, highlighting their unique characteristics and contributions to the field. We provide a table summarizing the characteristics of each dataset in Tab. \ref{tab:datasets}. For clarity, we categorize these datasets into two main types: real data and synthetic data, reflecting their origins and usage in the HTR field.

\subsubsection{Real data} 
\begin{figure*}
    \centering
    \includegraphics[width=\linewidth]{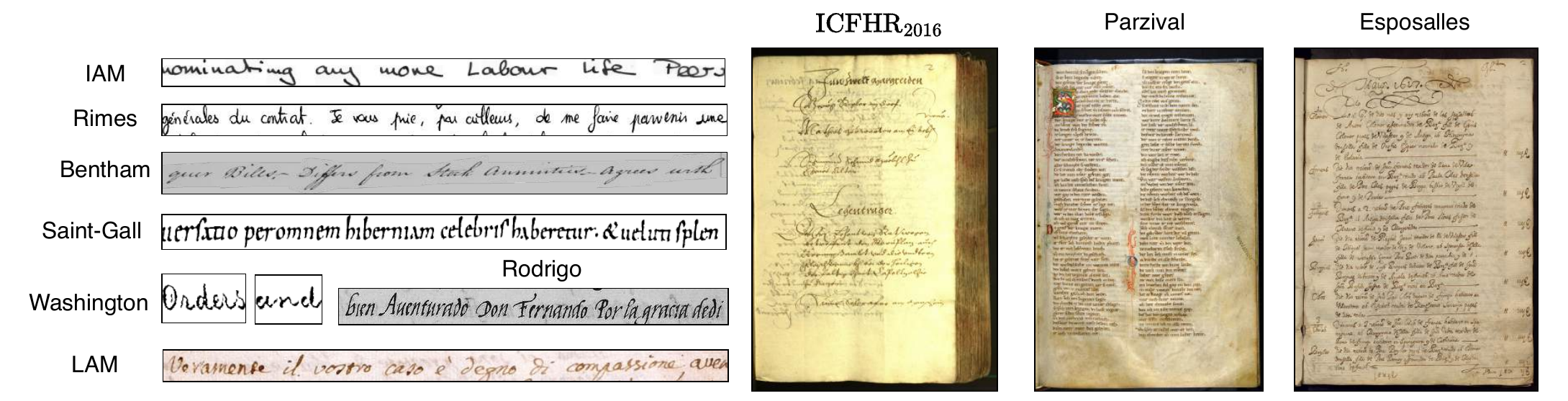}
    \caption{Examples of handwritten text samples from various datasets commonly used in HTR research. The image showcases a range of challenges, including diverse handwriting styles, languages (e.g., English, French, Latin, Spanish, and German), and document types, from clean modern handwriting to complex historical manuscripts. These variations highlight the complexity of tasks and the need for robust recognition methods.}
    \label{fig:real_datasets}
\end{figure*}

% The datasets used for handwritten text recognition (HTR) vary in language, time period, and annotation granularity, as summarized in Table~\ref{tab:datasets}. The IAM dataset (1999) remains the standard benchmark for English HTR, while Rimes (2011) is widely used for French. Historical datasets, such as Bentham (1748–1832), Saint-Gall (9th century), and Rodrigo (1545), provide valuable resources for studying older handwriting styles. Several datasets, including IAM, Rodrigo, and LAM, offer annotations at multiple levels (word, line, and page), supporting comprehensive training and evaluation. Others, such as Bozen and Parzival, focus on specific historical scripts, making them essential for research in medieval and historical text recognition. The Esposalles dataset (1451–1905) provides a unique collection of Catalan civil records, valuable for historical document processing. These datasets collectively cover a broad range of challenges, from modern handwriting variability to the complexities of ancient scripts.
We provide a comprehensive overview of widely used HTR datasets in Table~\ref{tab:datasets}, covering various languages, historical periods, and annotation levels. IAM~\cite{iamdatabase_marti_2002} remains the standard benchmark for English, while Rimes~\cite{rimes_2010} is commonly used for French. Historical datasets like Bentham \cite{bentham_causer2012building}, Saint-Gall \cite{saint_gall_scherrer1875verzeichniss}, and Rodrigo \cite{serrano-etal-2010-rodrigo} support research on ancient scripts. Bozen~\cite{icfhr_2016_sanchez} and Parzival~\cite{parzival_db} focus on medieval scripts. Esposalles~\cite{esposalles_2013} is particularly valuable for studying Catalan civil records.
These datasets, with their distinct languages, time periods, and annotation levels, provide diverse challenges and opportunities for advancing technologies across different historical and linguistic contexts. Examples of each real dataset can be seen in Fig. \ref{fig:real_datasets}.

\begin{table*}[h!]
\centering
\setlength{\tabcolsep}{5.2pt} % Increase column spacing
\caption{Datasets for the distinct levels at which HTR is performed.}
\begin{tabular}{@{}lcccccccccc@{}}
\toprule
Name              & Language        & Year              & Writers              & Word-level? & Line-level? & Page-level? & N. words    & N. lines & N. Pages  & Charset              \\ \midrule
IAM               & English         & $1999$              & $657$                  & \checkmark  & \checkmark  & \checkmark  & $115,520$     & $13,355$   & $1,539$     & $79$                   \\
Rimes             & French          & $2009$              & $1,300$                &   \checkmark  & \checkmark  &    \checkmark     &     $250,000$        &   $12,104$       & $12,723$    & $100$                  \\
Bentham (ICFHR$_{2014}$)          & English         & $1748$–$1832$         & $1$            &             & \checkmark  & \checkmark  &     $110,000$     &   $11,537$       & $433$    & $86$                   \\
Saint-Gall        & Latin           & $9^{th}$ cent.       & $1$                    & \checkmark  &  \checkmark   &             & $11,597$      & $1,410$    & $60$        & $49$                   \\
George Washington & English         & $1755$              & $2$                    & \checkmark  & \checkmark  &             &     $4,894$        &    $657$      & $20$        & $68$                   \\
Rodrigo           & Spanish         & $1545$              & $1$                    & \checkmark  & \checkmark  & \checkmark  & $232,000$     & $20,357$   & $855$       & $115$                  \\
Bozen (ICFHR$_{2016}$)        & German (mod) & $1470$–$1805$         & Unk.              &             &             & \checkmark  & $43,460$      & $10,550$   & $450$       & $92$                   \\
LAM               & Italian         & $1672$–$1750$           & $1$                    &  & \checkmark  & \checkmark  &        -     & $25,823$   & $1,171$     & $89$                   \\
Parzival               & German         & $1265$–$1300$           & $3$                    &  & \checkmark  & \checkmark  &  $23,478$    & $4,447$ & $45$    & $96$                   \\

Esposalles (total)               & Catalan         & $1451$–$1905$           & $1$                    & \checkmark  & \checkmark  &           & $67,311$   & $7,010$    & $202$ & $85$                  \\
\bottomrule
\end{tabular}
\label{tab:datasets}
\end{table*}
\renewcommand{\arraystretch}{1} % Reset to default after the table

\subsubsection{Synthetic data}
Synthetic data in HTR lacks a standardized dataset, leading each study to generate its own for pre-training. This has resulted in variations in text corpora, font choices, and dataset sizes, complicating direct comparisons across works.
For synthetic data, there are two factors that characterize the generation of these data, being the text that is rendered and the synthetic fonts used. In this section, we describe the different datasets used by the different works at the two levels proposed. The first synthetic dataset for HTR was the IIIT-HWS \cite{generating_krishan_2016}. % and used in \cite{improving_dutta_2018,gated_bluche_2017,word_krishnan_2018,deep_krishnan_2016,ganwriting_kang_2020,unsupervised_kang_2020,synthetic_etter_2019,candidate_kang_2019}, to name a few. %
The authors present the dataset containing 9M images as a result of rendering 90K words using the English dictionary Hunspell and 750 publicly available fonts. Kang et al. \cite{unsupervised_kang_2020} introduce a synthetic dataset leveraging 387 fonts and a 430,000 word multilingual corpus (English, French, Catalan and German) to generate a total of 5.6 million images containing also words.  %This is used in \cite{x,x,x,x,x}. 

The first synthetic dataset on the line level is found in \cite{pay_attention_kang_2022}. They rendered 138,000 handwritten lines from 387 fonts (probably the same as in \cite{unsupervised_kang_2020}) but using a text corpus from unspecified online e-books. Although initially set for full pages, Singh and Karayev generated text lines from WikiText-2 rendering 114 synthetic fonts \cite{Singh:ICDAR:2021}. In \cite{light_barrere_2022}, the authors use texts from English Wikipedia articles using online available handwritten fonts. They generate a total of 10K synthetic text-line images for each training epoch. The same authors also used texts from Wikipedia and historical books alongside 21 manually selected fonts to generate training data \cite{training_barrere_2024}. To pre-train a Transformer, in \cite{rescoring_wick_2021} 16M text lines from news, webpages and Wikipedia are rendered from ``several cursive computer fonts''. Using a very light CNN+Transformer encoder \cite{porwal_self-attention_2022}, the authors use the Brown corpus \cite{brown_corpus} and 400 handwriting-style fonts to render 50K lines of text. 
In TrOCR \cite{trocr_li_2023}, they pre-train a large Transformer with different sizes with 684M generic lines (first stage) + 17.9M specifically for handwritten text (second stage) using $5247$ synthetic fonts. For DTrOCR \cite{dtrocr_fujitake_2023}, they massively scale the pre-training of a decoder-only Transformer using 2B (2000M) lines and $5247$ synthetic fonts.  

For beyond-line level transcription, the first work that proposes synthetic generation was Singh et al.~\cite{Singh:ICDAR:2021}, where text was generated from the WikiText-2 dataset, as well as some images and mathematic formulae as artifacts to be detected by the system. The work of Coquenet et al.~\cite{Coquenet:TPAMI:2023b} composes synthetic images through the generation of lines with words from the IAM database, giving a more \textit{realistic} image output, sacrificing vocabulary richness. The work of Constum et al.~\cite{Constum:Arxiv:2024} acknowledges the limitations of this previous work---especially its tendency to overfit to this sinthetic data---and proposes an extension of the method from Singh et al., composing images from the WikiText-3~\cite{Merity:Arxiv:2016} dataset in three languages: English, french and german. Moreover, this method curates a selection of synthetic handwritten fonts and is specially programmed to mimic the layout structure found in mentioned datasets, such as Rimes or Bozen.     

\subsection{Performance assessment}
The assessment of transcription quality in {HTR} typically relies on computing edit distances. Edit distance, also known as Levenshtein distance, is a metric that quantifies how dissimilar two strings are. Specifically, it measures the number of insertions, deletions, or substitutions required to match a source string with a target one. The two well-known edit distance-based metrics are the \ac{CER} and the \ac{WER}. Let $\hat{y}$ be the hypothesis and $y$ be the ground truth sequence, and let us denote $L(\cdot, \cdot)$ as the function that calculates the Levenshtein distance between two character sequences. Then, these metrics are computed as:

\begin{equation}
    E(\hat{Y}, Y) = \frac{\sum_{i=1}^{n}L(y_{i}, \hat{y_i})}{\sum^{n}_{i=1}|y_i|}
\end{equation}

\noindent where $n$ is the number of samples in the test set and $|y_i|$ is the length of the sampled ground truth sequence in tokens. The only difference between \ac{CER} and \ac{WER} lies in the tokenization methods: characters or words, respectively. Lower values indicate better performance.

It is generally acknowledged that these metrics are suitable for up-to-line level transcription, since they effectively quantify character- and word-level errors, providing a direct measure of transcription accuracy in a constrained RO. However, as complexity increases, especially in document-level HTR, the system may make errors that are purely related to RO; for example, words that are correctly transcribed but not placed in their correct order according to the ground truth. In these situations, \ac{CER} and \ac{WER} may overly penalize such errors without providing insight into their cause.

This issue was firstly highlighted in the works of Antonacopoulos, Clausner and Pletschacher, where specific metrics---based on level of depth analysis---are proposed~\cite{Clausner:ICDAR:2013,Pletschacher2015}. These works, however, are not directly concerned with transcription quality evaluation. Some of them produce end-to-end results, while others deal with other related errors, such as LA.

An agnostic concept to solve the \ac{CER} and \ac{WER} strictness on alignment is the bag-of-words edit distance. This approach measures the error by comparing the frequencies of the terms of the vocabulary in both the transcript and the hypothesis. Thus, the problem is set up as a precision/recall measurement, one focusing on falsely predicted tokens and the other one on avoidance. This approach is found in several works of the literature~\cite{ClausnerPA20, Strobel2020, clausner2017a}, as well as it is a frequent benchmark in some competitions~\cite{ClausnerAP19,clausner2017a}. 

Another concept that has been adopted to assess the transcription quality of documents is the \textit{Hungarian Algorithm}, where the transcript words are matched with the ones in the ground truth that output the lowest edit distance. This concept has also been in several document recognition works~\cite{ClausnerPA20,tensmeyer2019training,long2022towards}. The most noteworthy approach is the ``Flexible Character Accuracy'' metric~\cite{ClausnerPA20, ClausnerAP19}.

Another idea that has been adopted in some works~\cite{sanchez:2017,wigington2018start,tensmeyer2019training} is to resort to the ``BLEU'' measure~\cite{papineni2002bleu}. This concept is borrowed from the Machine Translation field. It is based on matching the n-gram frequencies of the hypothesis with those of the reference sequence. Despite the avoidance of LA-related errors, this approach still suffers from the same problems that \ac{CER} and \ac{WER} have, which does not provide enough information about the source of the errors themselves.

Finally, the work of Vidal et al ~\cite{Vidal2023} proposes several modifications to the bag of words and the Hungarian Algorithm to measure adequately HTR beyond line-level transcription with regards to the RO. Specifically, they present, along with experimental insights, a set of three measures: traditional \ac{WER}, the \textit{h}\ac{WER}---a Hungarian Algorithm-based \ac{WER}---and $\Delta$\ac{WER}, which is the subtraction of the two previous metrics, which measures the proportion of error that is derived from RO misalignments in transcription. 

\begin{table}[ht!]
    \centering
    \renewcommand{\arraystretch}{1.0} % Stretch rows for better spacing
    \setlength{\tabcolsep}{1.2pt} % Increase column spacing
    \footnotesize % Adjust font size if needed
    \caption{Comparison of HTR models at the word and line level for the IAM database. `Par(M)' stands for million of parameters and `Alig.' for type of alignment. $\dagger$: Re-implementation by \cite{Cascianelli2022BoostingMA}.
    }
\begin{tabular}{@{}llccccc@{}}
\toprule
\multicolumn{1}{c}{\textbf{Model}} & \textbf{Year} & \textbf{Par(M)} & \textbf{Synth} & \textbf{LM/Lex.} & \textbf{Alig.} & \textbf{CER} \\ \midrule
MLSTM-CTC \cite{Graves2009_Line} & 2009 & 0.105 & - & Yes/30k & $\mathcal{L}_{ctc}$ & 18.2 \\
MLP/HMM \cite{improving_boquera_2011} & 2011 & - & - & Yes/20k & HMM & 9.8 \\
GMM/HMM \cite{hierarchical_dreuw_2011} & 2011 & - & - & Yes/50k & HMM & 10.1 \\
MLP/HMM \cite{hierarchical_dreuw_2011} & 2011 & - & - & Yes/50k & Yes & 12.4 \\
BLSTM-CTC \cite{liwicki_neural_2012} & 2012 & - & - & Yes/- & $\mathcal{L}_{ctc}$ & 18.2 \\
LSTM-HMM \cite{kozielski_improvements_2013} & 2013 & - & - & Yes/50k & HMM & 5.1 \\
GMM/HMM \cite{kozielski_open_2013} & 2013 & - & - & Yes/20k & HMM & 8.2 \\
LSTM-HMM \cite{doetsch_fast_2014} & 2014 & 10.7 & - & Yes/50k & HMM & 4.7 \\
MLSTM-CTC \cite{Pham_dropout_2014} & 2014 & 0.142 & - & Yes/50k & $\mathcal{L}_{ctc}$ & 5.1 \\
MLSTM-CTC \cite{bluche_deep_2015} & 2015 & 24 & - & Yes/50k & $\mathcal{L}_{ctc}$ & 4.4 \\
LSTM-HMM \cite{voigtlaender_sequence-discriminative_2015} & 2015 & 17 & - & Yes/- & HMM & 4.8 \\
MLSTM-CTC \cite{voigtlaender_handwriting_2016} & 2016 & 2.6 & - & Yes/- & $\mathcal{L}_{ctc}$ & 3.5 \\
MLSTM-Att. \cite{Bluche:NIPS:2016} & 2016 & - & - & Yes/50k & $\mathcal{L}_{ce}$ & 4.4 \\
C-BLSTM-CTC \cite{gated_bluche_2017} & 2017 & 0.725 & - & Yes/50k & $\mathcal{L}_{ctc}$ & 3.2 \\
C-LSTM-CTC \cite{multidimensional_recurrent_puig_2017} & 2017 & 9.3 & - & Yes/- & $\mathcal{L}_{ctc}$ & 4.4 \\
MLSTM-CTC \cite{Bluche:ICDAR:2017} & 2017 & - & - & - & $\mathcal{L}_{ctc}$ & 6.6 \\
MLSTM-CTC \cite{chen_simultaneous_2017} & 2017 & 0.686 & - & - & $\mathcal{L}_{ctc}$ & 11.1 \\
MLSTM-MLP/HMM \cite{castro_boosting_2018} & 2018 & 5.5 & - & Yes/50k & HMM & 3.6 \\
C-BLSTM-Att.-CTC \cite{gui_adaptive_2018} & 2018 & - & - & -/50k & $\mathcal{L}_{ctc}$ & 5.1 \\
C-BLSTM \cite{improving_dutta_2018} & 2018 & - & 90k-750f & -/90k & $\mathcal{L}_{ce}$ & 5.7 \\
C-BGRU-GRU \cite{kang_convolve_2018} & 2018 & - & - & - & $\mathcal{L}_{ce}$ & 6.8 \\
C-BLSTM-LSTM \cite{chowdhury2018efficient} & 2018 & 4.6 & - & - & $\mathcal{L}_{ce}$ & 8.1 \\
C-LSTM-Att. \cite{offline_sueiras_2018} & 2018 & - & - & -/50k & $\mathcal{L}_{ce}$ & 8.8 \\
C-LSTM-CTC \cite{word_krishnan_2018} & 2018 & - & Yes/90k & - & $\mathcal{L}_{ctc}$ & 9.8 \\
C-LSTM-Att. \cite{evaluating_michael_2019} & 2019 & 5 & - & - & Hybr. & 4.8 \\
C-BGRU-GRU-Att. \cite{candidate_kang_2019} & 2019 & - & 250k-387f & - & $\mathcal{L}_{ce}$ &  \\
Transformer \cite{pay_attention_kang_2022} & 2020 & 100 & 130k-387f & - & $\mathcal{L}_{ce}$ & 4.6 \\
Transformer \cite{pay_attention_kang_2022}$^\dagger$ & 2020 & 100 & 130k-387f & - & $\mathcal{L}_{ce}$ & 7.6 \\
C-CTC \cite{Yousef:CVPR:2020} & 2020 & 3.4 & - & - & $\mathcal{L}_{ctc}$ & 4.9 \\
C-CTC \cite{Huang2020} & 2020 & - & - & - & $\mathcal{L}_{ctc}$ & 6.1 \\
C-LSTM-Att. \cite{Coquenet:ICDAR:2021} & 2020 & 1.4 & - & Yes & $\mathcal{L}_{ce}$ & 7.7 \\
Transformer-CTC \cite{rethinking_diaz_2021} & 2021 & 8 & Yes & Yes/- & $\mathcal{L}_{ctc}$ & 2.8 \\
Transformer \cite{transformer_wick_2021} & 2021 & - & 1M- & - & $\mathcal{L}_{ce}$ & 5.67 \\
C-BLSTM-GRU-Att. \cite{characterbased_poulos_2021} & 2021 & - & - & - & $\mathcal{L}_{ce}$ & 16.6 \\
Transformer \cite{light_barrere_2022} & 2022 & 7.7 & Yes/- & - & Hybr. & 4.7 \\
Transformer Enc. \cite{porwal_self-attention_2022} & 2022 & 1.74 & 40k-400f & - & $\mathcal{L}_{ctc}$ & 7.5 \\
TrOCR (L) \cite{trocr_li_2023} & 2023 & 558 & 702M-5.4kf & 36-ch & $\mathcal{L}_{ce}$ & 2.9 \\
TrOCR (M) \cite{trocr_li_2023} & 2023 & 334 & 702M-5.4kf & 36-ch & $\mathcal{L}_{ce}$ & 3.4 \\
TrOCR (S) \cite{trocr_li_2023} & 2023 & 62 & 702M-5.4kf & 36-ch & $\mathcal{L}_{ce}$ & 4.2 \\
TrOCR (M) \cite{trocr_li_2023}$^\dagger$ & 2023 & 385 & 702M-5.4kf & - & $\mathcal{L}_{ce}$ & 7.3 \\
FCN-CTC (VAN) \cite{dan_coquenet_2023} & 2023 & 1.7 & - & - & $\mathcal{L}_{ctc}$ & 3 \\
DTrOCR \cite{dtrocr_fujitake_2023} & 2024 & 105 & 2B-5.4kf & 36-ch & $\mathcal{L}_{ce}$ & 2.4 \\
Transformer \cite{training_barrere_2024} & 2024 & 5.6 & 32M-21f & - & Hybr. & 4.2 \\
Transformer-CTC \cite{li_htr-vt_2024} & 2024 & 53.5 & - & - & $\mathcal{L}_{ctc}$ & 4.7 \\ \bottomrule
\end{tabular}
    \label{tab:line_methods}
\end{table}

\subsection{Performance comparison}
We compare state-of-the-art HTR methods across different levels of complexity, focusing on up-to-line transcription for words and lines, as well as beyond-line-level transcription for paragraphs and full documents. For the up-to-line level, we analyze the methodological evolution of these models, categorizing them based on their alignment strategies, such as CTC-based, sequence-to-sequence, and hybrid approaches, while also considering the impact of synthetic pre-training and external LMs. For beyond-line-level transcription, we explore the challenges posed by complex layouts and non-trivial RO, evaluating the effectiveness of segmentation-free models and unconstrained architectures. The comparison is structured around CER for up-to-line level (Table \ref{tab:line_methods}) and beyond-line level (Table \ref{tab:bll-results}). In both cases, we compare the results in terms of CER over the IAM dataset (for lines and paragraphs, respectively), as it is considered the standard benchmark for model comparison. Both tables report the architecture type along with the number of parameters (if reported in the original paper) and the amount of synthetic data used, expressed in terms of the number of examples and fonts employed. We report the CER results from the original papers for both levels. For up-to-line level transcription, we also report whether the method uses an LM and the vocabulary of the lexicon.

Early methods exhibited a high CER of up to 18.2 \% \cite{Graves2009_Line}. However, these errors were significantly reduced over a span of just 4--5 years, reaching as low as 5.1 CER points \cite{Pham_dropout_2014, bluche_deep_2015} or even lower \cite{voigtlaender_handwriting_2016, deep_wicht_2016}. This significant improvement can largely be attributed to the integration of LMs and lexicons, which substantially facilitated the recognition process, alongside the increasing size and fine-tuning capacity of HTR models. In subsequent years, methodological advancements have focused on maintaining or surpassing these results without relying on LM and lexicons. This has come at the cost of a considerable increase in model parameters \cite{pay_attention_kang_2022, trocr_li_2023} and computational expense, as well as the incorporation of large pre-training with synthetic data \cite{trocr_li_2023, dtrocr_fujitake_2023}. These trends reflect the ongoing evolution of HTR, as discussed in the following sections.

\begin{table}[]
\centering
\caption{Comparison of HTR models beyond line-level for the IAM database. The `Type' column represents the categories in Section \ref{sec:bll_description} (MSK: masking; UNF: unfolding; UNC: unconstrained).}

\label{tab:bll-results}
\begin{tabular}{@{}lccccc@{}}
\toprule
\multicolumn{1}{c}{Model} & Year & N. Params (M) & Synth & Type & CER \\ \midrule
JLSAT \cite{Bluche:NIPS:2016} & 2016 & - & - & MSK & 6.8 \\
SAAR \cite{Bluche:ICDAR:2017} & 2017 & - & - & MSK & 16.2 \\
Origaminet \cite{Yousef:CVPR:2020} & 2020 & 14.0 & - & UNF & 4.7 \\
SPAN \cite{Coquenet:ICDAR:2021} & 2021 & 19.2 & - & UNF & 5.4 \\
FPHR \cite{Singh:ICDAR:2021} & 2021 & 27.9 & 58B-34f & UNC & 6.3 \\
TBPHTR \cite{Rouhou:PRL:2022} & 2022 & 26.9 & - & UNC & - \\
VAN \cite{Coquenet:TPAMI:2023} & 2023 & 11.0 & - & MSK & 4.6 \\
DAN \cite{Coquenet:TPAMI:2023b} & 2023 & 18.6 & 13k-0f & UNC & - \\
MSDocTr-Lite \cite{Dhiaf:PRL:2023} & 2023 & 26.9 & - & UNC & 6.4 \\ \bottomrule
\end{tabular}
\end{table}

In the case of beyond-line-level approaches, we observe a significant evolution in recognition accuracy. Early methods, such as the \ac{SAAR}, initially reported high error rates, 16.2\%.\footnote{While \ac{JLSAT} is historically recorded as the first beyond line-level method, evidence suggests that \ac{SAAR} was developed earlier but published later due to publication delays.} The integration of the CTC loss function marked a substantial improvement in the paradigm, as first demonstrated by the \ac{JLSAT}, which reduced the error rate by 58\%. The CTC approach continues to demonstrate superior performance for beyond line-level IAM. Current state-of-the-art models, the \ac{VAN} and Origaminet, exhibit error rates of 4.7 and 4.6\% respectively. These results contrast with the best-performant unconstrained approach, the MsDocTr-Lite, which yields a 6.4\% \ac{CER}. However, the specific approach to addressing this challenge appears less critical, as performance remains tightly clustered across methods. This suggests that the key factor lies in adapting the document into a line-level structure for transcription using traditional methods.

It is important to note that this comparison has limitations, as the IAM dataset only represents difficulty up to the paragraph level. This explains why models like the DAN, which reports state-of-the-art transcription performance, opt to exclude this dataset in favor of corpora such as Rimes or Bozen. These models excel at handling more complex difficulties, while paragraph-level recognition can be effectively addressed through line-level adaptation techniques, such as masking or unfolding, combined with CTC-based transcription.

% ===================
% === CONCLUSIONS ===
% ===================
\section{Conclusions, challenges \& future directions}
\label{sec:conclusions}
In this work, we examined the methodological evolution of the field of Handwritten Text Recognition (HTR). We traced its progression from early handcrafted approaches relying on heuristics to transcribe individual words to modern systems capable of automatically transcribing entire documents. To structure our analysis, we divided the study into two main methodological categories: approaches that focus on transcription at the line level (words and lines) and those that extend beyond the line level (paragraphs, pages, and full documents). Additionally, we reviewed the most commonly used datasets in the literature and the evaluation methods applied to these systems. While the field has made significant advancements, there remains considerable room for improvement. In the following sections, we address the current challenges faced by the field and propose potential future directions for its development.

\subsection{Common comparison framework}
A significant challenge in HTR research is the lack of a unified comparison framework, leading to three key issues: reliance on single datasets, inconsistencies in character set tokenization, and limited reproducibility. Many studies evaluate models only on datasets like IAM or Rimes, restricting generalizability and failing to assess robustness across diverse handwriting styles and languages. Furthermore, the use of synthetic data has begun to be the standard, making unfair comparisons. Additionally, advancements in tokenization, influenced by LLMs, have introduced disparities in character set sizes, with some models handling large charsets (e.g., multilingual alphabets) while others operate on smaller sets \cite{trocr_li_2023, dtrocr_fujitake_2023}, creating unfair comparisons due to differing task complexities. Compounding these problems, many papers do not provide code or sufficient implementation details, hindering reproducibility and reliable benchmarking. To address these issues, the field must prioritize cross-dataset evaluations, standardize tokenization practices, encourage code sharing, and adopt multilingual and domain-diverse benchmarks to enable fair and meaningful comparisons. 

% \begin{figure}[h]
%     \centering
%     \includegraphics[width=\columnwidth]{figures/cer_evolution.pdf}
%     \caption{Evolution of CERs on the IAM dataset at the line-level throughout the years. The size of the circle represents the number of parameters (M).}
%     \label{fig:evolution_cer}
% \end{figure}

\subsection{Generalization \& Evaluation}
In the early stages of HTR research, the focus in the literature was primarily on improving recognition accuracy on test splits from established competitions, measuring in-distribution (ID) generalization. As models have grown more capable of fitting the training distribution, the error rates on these test partitions have steadily decreased, as illustrated in Table \ref{tab:line_methods}. However, this reduction in error often comes at the cost of overfitting to the specific test sets of datasets like IAM or Rimes. In this context, we argue that future benchmarks should prioritize evaluating performance on out-of-distribution (OOD) data to better assess models' capabilities in highly challenging scenarios. Specifically, models should demonstrate the ability to perform reasonably well in cross-lingual environments, where the training and test sets are in entirely different languages. While there has been an initial exploration of these generalization capabilities in \cite{garrido2024_generalization}, this study is limited to the HTR task at the line level, leaving broader aspects of generalization unexplored.

\subsection{Fine-tuning \& Adaptation}
Many of the presented systems encounter difficulties when dealing with limited data \cite{boosting_aradillas_2021, wigington2018start}, which is often the case with historical sources. Most of the methods---specially the most modern ones--- rely on synthetic data generation \cite{light_barrere_2022, training_barrere_2024, trocr_li_2023, dtrocr_fujitake_2023}, as it is a general-purpose solution that bypasses this obstacle and, sometimes, provides more general solutions. That is, most of the state-of-the-art are pre-trained models with synthetic data that are then fully-fine-tuned with the target corpora. Despite being proven an effective method, full fine-tuning may not be the most convenient technique for this task. There exist many specific methods that may help fine-tuned HTR models to obtain better performance with few data \cite{castro_boosting_2018, boosting_aradillas_2021, finetuning_koht_2023, unsupervised_kang_2020, metahtr_bhunia_2021}.

\subsection{Future directions}
Future directions in HTR may increasingly explore the integration of Vision-Language Models (VLM) \cite{Beyer2024PaliGemmaAV, Liu2023VisualITLlava}, driven by advancements in computational capabilities and the availability of large-scale synthetic corpora, building on their successful application in Optical Character Recognition (OCR) \cite{LayoutLM_2020, Kim2021OCRFreeDUDonut}. VLMs offer the potential to unify visual and linguistic information, which could enhance generalization across diverse handwriting styles and scripts. However, it remains uncertain how these models will be optimized for efficiency and real-world applicability, particularly given the high computational costs associated with large-scale training. Additionally, while synthetic data plays a crucial role in adapting real HTR datasets and improving performance \cite{trocr_li_2023, dtrocr_fujitake_2023}, its effectiveness in replicating real-world variability is still an open question. Future research will likely focus on assessing the feasibility and limitations of VLM in HTR, particularly for historical manuscripts and low-resource languages, where traditional methods still play a dominant role.

On the other hand, Self-Supervised Learning (SSL) represents one of the most recent and competitive paradigms within the deep learning field, aimed at alleviating the large amount of labeled data required by deep neural models for learning \cite{LeCunn2021}. While traditional supervised learning relies on human-annotated corpora, SSL is designed to learn through pseudo-labeled data—where no human annotation is involved—before being applied to one or more downstream tasks (e.g., classification) \cite{Ozbulak2023}. Currently, this paradigm represents an effective solution to data-limited scenarios, providing multipurpose state-of-the-art models \cite{Delvin2019, Bao2021}. SSL has already been tested in up-to-line transcription scenarios with remarkable results \cite{Aviad2021}, although it has not been fully explored. It could represent a breakthrough in HTR, enabling training on real data without the need for synthetic datasets. Some adaptations could be proposed for image-to-sequence approaches---similar to how BEiT \cite{Bao2021} is structured---as they better fit the classification pretext task.

%============== 
 
\section*{Acknowledgments}
The first author is supported by grant CIACIF/2021/465 from ``Programa I+D+i de la Generalitat Valenciana''. The second autor is supported by grant ACIF/2021/356 from the ``Programa I+D+i de la Generalitat Valenciana''.


%\section{References}
\bibliographystyle{IEEEtran}
\bibliography{IEEEabrv,bibliography.bib}

% your own custom macro containing the $\backslash${\tt{includegraphics}} command to make things
 

%\bf{If you include a photo:}\vspace{-33pt}
%\begin{IEEEbiography}[{\includegraphics[width=1in,height=1.25in,clip,keepaspectratio]{fig1}}]{Michael Shell}
%Use $\backslash${\tt{begin\{IEEEbiography\}}} and then for the 1st argument use $\backslash${\tt{includegraphics}} to declare and link the author photo.
%Use the author name as the 3rd argument followed by the biography text.
%\end{IEEEbiography}



%\vfill

\end{document}